\documentclass[10pt, journal]{IEEEtran}

\usepackage{booktabs} 

\usepackage{url}

\usepackage[numbers]{natbib} 

\usepackage[ruled]{algorithm2e} 

\SetAlFnt{\small}
\SetAlCapFnt{\small}
\SetAlCapNameFnt{\small}
\SetAlCapHSkip{0pt}
\IncMargin{-\parindent}

\makeatletter
\newcommand{\removelatexerror}{\let\@latex@error\@gobble}
\makeatother

\usepackage[export]{adjustbox}
\usepackage{amssymb}
\usepackage{amsthm}
\usepackage{amsmath}
\usepackage{comment}
\usepackage{algorithmic}
\usepackage{amsmath}
\usepackage{moresize}
\usepackage{helvet}
\usepackage{courier}
\usepackage{graphicx}
\usepackage{wrapfig}
\usepackage{url}
\usepackage[font=small,labelfont=bf]{caption}
\usepackage{subcaption}
\usepackage{varwidth}

\usepackage[utf8]{inputenc} 
\usepackage[T1]{fontenc}    
\usepackage{url}            
\usepackage{booktabs}       
\usepackage{amsfonts}       
\usepackage{nicefrac}       
\usepackage{microtype}      
\usepackage{multirow}       
\def\i#1{\hbox{\it #1\/}}
\def\beq{\begin{equation}}
\def\eeq#1{\label{#1}\end{equation}}
\def\ba{\begin{array}}
\def\ea{\end{array}}

\def\t{\texttt{True}}
\def\f{\texttt{False}}
\def\iif{\hbox{\bf if}}

\def\causes{\hbox{\bf causes}}
\def\inertial{\hbox{\bf inertial}}
\def\default{\hbox{\bf default}}

\def\nonex{\hbox{\bf nonexecutable}}
\def\Epsilon{\text{E}}

\usepackage{xcolor}

\newtheorem{theorem}{Theorem}

\begin{document}
\title{TDM: Trustworthy Decision-Making via Interpretability Enhancement}

\author{Daoming~Lyu,
        Fangkai~Yang,
        Hugh~Kwon,
        Wen~Dong, \\
        Levent~Yilmaz,~\IEEEmembership{Member,~IEEE,}
        and~Bo~Liu,~\IEEEmembership{Senior Member,~IEEE}
\thanks{\copyright 2021 IEEE.  Personal use of this material is permitted.
Permission from IEEE must be obtained for all other uses, in any current 
or future media, including reprinting/republishing this material for 
advertising or promotional purposes, creating new collective works, for 
resale or redistribution to servers or lists, or reuse of any copyrighted 
component of this work in other works.}
\thanks{D. Lyu, H. Kwon, L. Yilmaz, and B. Liu are with the Department
of Computer Science and Software Engineering, Auburn University, Auburn,
AL, 36849 USA. 
F. Yang is with NVIDIA Corporation, Bellevue, WA, 36849, USA.
W. Dong is with the Department
	of Computer Science and Engineering, University at Buffalo, Buffalo,
	NY, 14260 USA.
}
\thanks{Corresponding author: Bo Liu~$<$boliu@auburn.edu$>$.}}

\markboth{IEEE Transactions on Emerging Topics in Computational Intelligence (TETCI),~Vol.~00, No.~0, Month~0000}%
{TDM: Trustworthy Decision-Making via Interpretability Enhancement}

\maketitle

\begin{abstract}
Human-robot interactive decision-making is increasingly becoming ubiquitous, and trust is an influential factor in determining the reliance on autonomy. However, it is not reasonable to trust systems that are beyond our comprehension, and typical machine learning and data-driven decision-making are black-box paradigms that impede interpretability. Therefore, it is critical to establish computational trustworthy decision-making mechanisms enhanced by interpretability-aware strategies.
To this end, we propose a Trustworthy Decision-Making (TDM) framework, which integrates symbolic planning into sequential decision-making. The framework learns interpretable subtasks that result in a complex, higher-level composite task that can be formally evaluated using the proposed trust metric. TDM enables the subtask-level interpretability by design and converges to an optimal symbolic plan from the learned subtasks. Moreover, a TDM-based algorithm is introduced to demonstrate the unification of symbolic planning with other sequential-decision making algorithms, reaping the benefits of both. Experimental results validate the effectiveness of trust-score-based planning while improving the interpretability of subtasks.
\end{abstract}


\section{Introduction}

\IEEEPARstart{F}{rom} self-driving cars to voice assistant on the phone, artificial intelligence (AI) is progressing rapidly. Though AI systems are undeniably powerful and continue to expand their role in our daily lives, emerging issues, such as lack of transparency and uncontrolled risk in decision-making, give rise to a vital concern:\textbf{\textit{ why should the end-users trust the decision support system?}} Reliance on technology and autonomy is strongly influenced by trust \cite{lee2004trust}.
However, a recent study reveals that a mere belief in interacting with autonomous teammate led to diminished performance and passive behavior among human participants even though they were actually collaborating with a remote human partner \cite{demir2018impact}, indicating that we do not trust the capabilities of AI systems as much as we trust humans.

Building trustworthy AI systems 
\cite{marsh1992trust, o2005trust, ribeiro2016should, ong2019air}
, therefore, becomes an important issue for humans to reap the full spectrum of societal and industrial benefits from AI, as well as for ensuring safety in the use of the system \cite{hoff2015trust}.
Indeed, in modern artificial intelligence systems, there are many interactive relations, primarily in the form of \textit{interactions} between humans, between humans and the smart agents (human-computer interaction systems), between agents and the environment (interactive machine learning), and among smart agents (multi-agent systems). 

It is also important to note that the trust is mutual. An AI agent must also gauge the trustworthiness of the end-users. However, very few studies exist in agent-to-human trust models. The recent Boeing 737 MAX 8 incidents \cite{josephs2019boeing, Ortiz2019boeing} have tragically shown the dangers of the lack of the automated agent's ability to trust the end-users.
The pilots of the crashed airplanes attempted to take control back from a malfunctioning flight control system, but the flight control system could not respond to the change of trust from pilots. Instead of trusting the pilots' decisions, the flight control system continued to override the manual control resulting in fatal crashes.

Besides, AI systems, in general, need to comply with the norms of the social and cultural context if they are expected to operate side by side with humans. Therefore, trust is particularly important for normative agents that are expected to exhibit behavior consistent with the norms of a society or group. From the humans' perspective, we trust things that behave as we expect them to, indicating that trust derives from the process of minimizing the perceived risk~\cite{rousseau1998not}. Perceived risk is often defined in terms of uncertainty about the possibility of failure or the likelihood of exhibiting improper behavior.

The typical approach to computationally modeling trust is through beliefs. In multi-agent settings, the agents maintain beliefs on the trustworthiness of other agents influenced by mutual interactions \cite{castelfranchi2003trust}.
When end-users are involved, the agent maintains beliefs about the \emph{beliefs} of humans', concerning the trustworthiness of the agent \cite{chen2018trust}.
It is difficult to design an explicit measure of trust without considering different perspectives.
Recent work of \citeauthor{hind2018increasing}~\cite{hind2018increasing} suggested several elements, such as fairness, robustness, and explainability, to increase trust in AI systems.

To open up the AI black-box and facilitate trust, it is critical to enable the AI system to be reliable, transparent, and explainable so that the system can achieve reproducible results, justify the decisions it makes, and understand what is important in the decision-making process. And this combination of features can be called ``interpretability''.
Deep learning algorithms have been successfully applied to sequential decision-making problems involving high-dimensional sensory inputs such as Atari games~\cite{dqn:nature:2015}.
However, deep learning
algorithms have limited interpretability and transparency, which gives rise to doubt by the human end-users due to a lack of trust and confidence in the AI systems.
Prior research efforts \cite{biran2017human} on interpretability often focus on explaining the results of learning when the underlying problem and architecture are complex. Symbolic planning \cite{hanheide2015robot,chen2016planning,khandelwal2017bwibots, jeong2016task, lyu2019sdrl} has been used to break down the complex task into simpler interpretable subtasks.
However, existing studies \cite{leonetti2016synthesis, yang:peorl:2018} focus on the performance gain of symbolic planning when it is applied to general sequential decision-making rather than interpretability.

Motivated by understanding the task-level behavior of the algorithm and elevating the trust between human end-users and the smart agents, we propose a unified framework of Trustworthy Decision-Making.
Symbolic planning is utilized to perform reasoning and planning on explicitly represented knowledge, which results in task-level interpretability.
We also introduce a computational trust metric based on the success ratio of symbolic tasks. The key insight is that the success is better measured when the tasks are interpretable. Achieving a reliable success rate of the tasks is equivalent to minimizing perceived risks of the agent's behavior, thereby enhancing its trustworthiness. 

The rest of the paper is structured as follows. Section~\ref{sec:related} presents related research in trust, interpretability, followed by a review of prior trust evaluation studies in the extant literature from different perspectives. Section~\ref{sec:prelim} introduces the background on symbolic planning along with an overview of the interactive sequential decision problem. Section~\ref{sec:sdqnim} provides a detailed description of the TDM framework.
We introduce a learning algorithm facilitated by our formal framework in Section~\ref{sec:alg}. The optimality analysis for the converged plan generated by the algorithm is presented to substantiate the utility of our strategy.
The experimental results of Section~\ref{sec:exp} validate the interpretability of the high-level behavior of the system and the effectiveness of the trust metric while maintaining improved data efficiency.
Section~\ref{sec:conclusion} concludes by summarizing our results and delineating potential avenues of future research.


\section{Backgrounds}\label{sec:related}

This section identifies and elaborates several key components of a computational framework for trustworthy decision-making, i.e., the definition of trust, interpretability and its relation to trust, and computational models of trust metrics.

\noindent\textbf{Trust}
Trust is a subjective perception affected by many factors. One essential impact factor is risk \cite{rousseau1998not}. Following this direction, \citeauthor{marsh1992trust}~\cite{marsh1992trust, marsh1994formalising} presented a formalization of trust and its uses in distributed AI; \citeauthor{liu2018block}~\cite{liu2018block} proposed a stochastic block coordinate ascent policy search algorithm to address the risk management in dynamic decision-making problems. From the perspective of human social interactions, \citeauthor{lee2004trust}~\cite{lee2004trust} studied trust in automation. Specifically, in the context of AI, \citeauthor{o2005trust}~\cite{o2005trust} presented two computational models of trust for the recommender systems; \citeauthor{ribeiro2016should}~\cite{ribeiro2016should} proposed the Local Interpretable Model-agnostic Explanations framework to explain the predictions of any classifier so that the model can be transformed into a trustworthy one. In addition, \citeauthor{hind2018increasing}~\cite{hind2018increasing} proposed the supplier's declaration of conformity to increase trust in AI services.
Another important factor of trust is performance. Prior research in robotics suggests performance and risk are related factors. \citeauthor{desai2013impact}~\cite{desai2013impact} demonstrated how a robot's failure influences human trust in real-time. \citeauthor{chen2018trust}~\cite{chen2018trust} proposed to use deliberate failures to calibrate the trust level. In this work, the human users perceive potential failures as risks and then respond according to the level of trust they place in the robot's ability.

\noindent\textbf{Interpretability} 
A key component of an AI system is the ability to explain its decisions, recommendations, predictions, or actions along with the process through which they are made. Studies on interpretability focus on describing the internals of a system in an understandable way to humans \cite{lipton2016mythos, doshi2017towards, gilpin2018explaining}. 
\citeauthor{biran2017human}~\cite{biran2017human} proposed an approach to justify the prediction to the user rather than only explaining how the prediction is reached automatically. Another line of research attempts to explain the Markov Decision Processes (MDPs). \citeauthor{elizalde2007mdp}~\cite{elizalde2007mdp} devised an intelligent assistant to assist the power plant operator, which can explain commands generated by an MDP-based planning system. \citeauthor{khan2009minimal}~\cite{khan2009minimal} explored the minimal sufficient explanation of policies for factored MDP by populating a set of domain-independent templates.
Symbolic planning has been integrated with reinforcement learning in the previous attempts \cite{leonetti2016synthesis, yang:peorl:2018} when the underlying MDP is relatively simple.
Deep reinforcement learning is often used for larger MDP, and in this direction, the program induction approach is used instead to enable policy interpretability \cite{verma2018programmatically}.
Our work admits the symbolic knowledge to enable task-level interpretability for sequential decision-making with function approximation, such as deep reinforcement learning, and it is interpretable by construction.

\noindent\textbf{Computational Models of Trust Metric}
Studies have been conducted on how to evaluate trust quantitatively. \citeauthor{schmidt2007fuzzy}~\cite{schmidt2007fuzzy} proposed a customized trust evaluation model based on fuzzy logic for multi-agent systems. \citeauthor{huynh2006integrated}~\cite{huynh2006integrated} presented a decentralized model to evaluate trust in open systems. \citeauthor{castelfranchi2003trust}~\cite{castelfranchi2003trust} devised a socio-cognitive model of trust by using the fuzzy cognitive maps and introduced a degree of trust derived from the credibility of the trust beliefs. Also, there exists research about trust evaluation in the context of computer networks. \citeauthor{theodorakopoulos2004trust}~\cite{theodorakopoulos2004trust} focused on the evaluation processes of trust evidence in Ad Hoc Networks. \citeauthor{sun2005trust}~\cite{sun2005trust} presented an information-theoretic framework to quantitatively measure trust and model trust propagation in Ad Hoc Networks. \citeauthor{yan2003trust}~\cite{yan2003trust} and \citeauthor{sun2006trust}~\cite{sun2006trust} both proposed trust evaluation frameworks concerning  security in computer networks.
In the context of HRI, one of the earliest trust evaluations is proposed by \citeauthor{lee1992trust}~\cite{lee1992trust} and expressed as a regression series.
\citeauthor{xu2016towards}~\cite{xu2016towards} also used a regression series to evaluate trust in autonomous vehicles. In another work, \citeauthor{xu2015optimo}~\cite{xu2015optimo} proposed a Bayesian model.
Trust has also been evaluated as states of MDP \cite{chen2018trust}.

To the best of our knowledge, there exists limited and inconclusive research on establishing a computational model of trust to measure the quality of a proposed task. In a quality-centric perspective, the quality of behavior should correspond to the plan's trust score that indicates the level of trustworthiness.
Such metrics need to be general enough to make evaluations across different tasks. A relevant research area in psychology is called intrinsic motivation, which is defined as the level of inherent satisfaction by accomplishing an activity. 
Interpersonal trust is strongly correlated to the trustor's perceived intrinsic motivation of trustee \cite{holmes1985interpersonal_trust,moorman1993interpersonal_trust}. Combined with the role of interpretability in trustworthiness, this means a human is more likely to develop feelings of trust towards a smart agent if the human understands that the agent is intrinsically motivated to improve its abilities.
    
Different from extrinsic motivation, the inherent satisfaction is driven by an internal utility function \cite{oudeyer2009intrinsic}. This makes it a suitable general metric to make evaluations of different tasks. Intrinsically-motivated learning \cite{intrinsic:barto:2005} used the framework of options.
An option, or a subtask, can be initiated in some states to control the agent's behavior in a subset of the environment with a terminating condition.
Motivated by these factors, we propose a trust metric evaluation framework based on the internal utility function, which can be used to compute the trustworthiness values for the trust score and evaluate the high-level behavior of the system. One of the problems of interpretability is that it can induce unwarranted trust of unreliable agents \cite{wang2016transparency,yang2017transparency}. We hypothesize that the quality-based metric will alleviate this problem by justifying why the learned plans are good and trustworthy, whereas the abandoned plans are not, thus ensuring the agent's learning process is reliable.


\section{Preliminaries}\label{sec:prelim}

In this section, we introduce the motivation for developing a trust metric and the formulation of a symbolic planning strategy with respect to sequential decision making.

\noindent\textbf{Interactive Sequential Decision-Making.}
In sequential decision-making problems, an agent takes a sequence of actions to determine its utility. However, the utility is often difficult to model in practical settings when the environment is complex or changes dynamically. In such cases, the utility is determined through feedback from human interaction. We will demonstrate an environment that changes over time in our experiments but will simplify it to model the utility changes to allow agents to train quickly without interaction. One approach is to design multiple similar tasks in differing complexities, learn simple tasks first, and transfer the knowledge to learn more complex tasks \cite{peng2018curriculum}. Instead, we will break down complex tasks into simpler sub-tasks.
And since it is expensive to consider a complete sequence of decisions, we will model our problem with Markovian property.

Consider a Markov Decision Process (MDP)
\footnote{We follow the style of notation in the 1st edition of reinforcement learning book by \citeauthor{rlbook:sutton2007:1stedition}~\cite{rlbook:sutton2007:1stedition}.}
defined by a tuple $({\mathcal{S},\mathcal{A},P_{ss'}^{a},r,\gamma})$.
We will use the MDP to model the sequential decision-making problem in symbolic representation, where $\mathcal{S}$ is the set of symbolic states and $\mathcal{A}$ is the set of actions.
Then $P_{ss'}^{a}$ is the transition kernel that, given a state $s\in\mathcal{S}$ and an action $a\in\mathcal{A}$, defines the probability the next state will be $s'\in\mathcal{S}$.
Moreover, $r(s,a):\mathcal{S}\times\mathcal{A}\mapsto\mathbb{R}$ is a reward function bounded by $r_{\max}$, and $0\leq\gamma<1$ is a discount factor. A solution to an MDP is a policy $\pi:\mathcal{S}\mapsto \mathcal{A}$ that maps a state to an action.

To evaluate a policy $\pi$, there are two types of performance measures: the expected discounted sum of reward for infinite-horizon problems 
and the expected un-discounted sum of rewards for finite horizon problems. In this paper we adopt the latter metric defined as $J^\pi_{\rm avg}(s) = \mathbb{E}[\sum\limits_{t = 0}^T {{r_t}}|s_0=s ]$. We define the \textit{gain reward} ${\rho ^\pi }(s)$ reaped by policy $\pi$ from $s$ as
$
{\small
{\rho ^\pi }(s) = \mathop {\lim }\limits_{T \to \infty } \frac{{J^\pi_{{\rm{avg}}}(s)}}{T} = \mathop {\lim }\limits_{T \to \infty } \frac{1}{T}\mathbb{E}[\sum\limits_{t = 0}^T {{r_t}} ]
} .
$
Gain reward ${\rho ^\pi }(s)$ is often used to measure the lower bound of the expected cumulative rewards of a task w.r.t a given policy. For example, ${\rho ^\pi }(s)$ is instrumental in deciding whether a specific task is rewarding enough to make the ``stay-or-leave'' decision~\cite{shuvaev2020r}. More mathematical details of gain reward can be referred to~\citep{puterman,averageRL:mahadevan1996average,tadepalli1994h,mahadevan2010basis}.

\noindent\textbf{Symbolic Planning} (SP) has been used in tasks that must be highly interpretable to human users, such as interacting and cooperating with mobile robots \cite{hanheide2015robot,chen2016planning,khandelwal2017bwibots, jeong2016task, lyu2019sdrl}.
By abstracting the planning problem as symbols, SP can solve the problem based on logical reasoning.
A symbolic representation is constructed so that it contains the knowledge of objects, properties, and the effects of executing actions in a dynamic system.
Such representation can be implemented via a formal, logic-based language such as the Planning Domain Definition Language (PDDL) \cite{mcdermott1998pddl} or an action language \cite{gel98} that relates to logic programming under answer set semantics (answer set programming) \cite{lif08}.

SP algorithms are also white-box algorithms.
With the symbolic knowledge designed to be human-readable, the behavior of an SP agent that plans and reasons based on it is naturally interpretable.
Prior works combine symbolic planning with other sequential-decision making algorithms to bring stability to plan-based learning \cite{leonetti2016synthesis, yang:peorl:2018}.
However, these works only highlight the performance improvement from the prior domain knowledge represented by SP.

\noindent\textbf{Action Language $\mathcal{BC}$.} An {\em action description}~$D$ in the language $\mathcal{BC}$ \cite{lee13} includes two kinds of symbols on signature $\sigma$, {\em fluent constants} that represent the properties of the world, and {\em action constants}
, representing actions that influences the world.
A \textit{fluent atom} associates a fluent constant $f$ to a value $v$ and is expressed as $f=v$. In particular, we consider a boolean domain for $f$.
{\em Causal laws} can now be defined on the fluent atoms and action constants, describing the relationship among fluent atoms and the effects of actions on the value of fluent atoms.
A {\em static law} may state that a fluent atom $A$ is true at a given time step when $A_1,\ldots, A_m$ are true
$
(A~\iif~A_1,\ldots,A_m)
$
or
the value of $f$ equals~$v$ by default
$
(\default~f=v)
$.
On the other hand, a {\em dynamic law} describes an action $a$
$
(\nonex~a~\iif~A_1,\ldots,A_m)
$
or the effect of $a$ on the fluent atom $A$
$
(a~\causes~A~\iif~A_1,\ldots, A_m)
$.
An inertia is a special case of dynamic law that states the value of fluent constant $f$ does not change with time
$
(\inertial~f)
$.
An action description is a finite set of causal laws, capturing the domain dynamics as transitions.

Therefore, given action description~$D$, here is an example of how to perform symbolic planning with action language $\mathcal{BC}$. 
Let	$\langle s,a,s' \rangle$ denote a transition from a symbolic state $s$ to a symbolic state $s'$ by a set of action $a$.
Then a planning problem can be formulated as a tuple $(I, G, D)$. The planning problem has a plan of length $l-1$ if and only if there exists a transition path of length $l$ such that $I=s_1$ and $G=s_l$. In the rest of the paper, $\Pi$ denotes both the plan and the transition path by following that plan. Given the above semantics represented in $\mathcal{BC}$, automated planning can be achieved by an answer set solver such as \textsc{Clingo}~\cite{gekasc12c}, and provide the solution to the planning problem.


\section{The TDM Framework}
\label{sec:sdqnim}
We will also use the MDP to model the underlying sequential decision-making problem, define by a tuple $(\widetilde{\mathcal{S}},\widetilde{\mathcal{A}},\widetilde{P^a_{ss'}},r,\tilde{\gamma})$,
where $\widetilde{\mathcal{S}}$ consists of states of high-dimensional sensory inputs such as pixel images, $\widetilde{\mathcal{A}}$ is the set of primitive actions,
and others are similarly defined as before.
The goal is to learn both a sequence of subtasks and the corresponding sub-policies, so that executing the sub-policy for each subtask one by one can achieve maximal cumulative reward.
It should be noted that our framework is not restricted to high-dimensional sensory inputs. However, we choose to focus on them since learning from such inputs is more challenging and relevant to real-world applications.

We assume human experts can provide a symbolic structure of the problem.
The symbolic structure consists of the knowledge of objects, properties, and the preconditions and effects of executing subtasks in a given problem domain.
Although this appears to be a lot of effort, it is possible to develop general-purpose action modules \cite{erdo06,erdo08,inclezan2016modular} as it has been shown that dynamic domains share many actions in common.
Therefore, the symbolic formulation for one problem can be adapted to another with a little effort by instantiating a different set of objects or adding a few more rules.
Our framework's domain dynamics have a coarse granularity and high-level abstraction, enabling the decision-making process to be robust and flexible when facing uncertainty and domain changes.

\begin{figure}[!t]
\centering
\includegraphics[height=4cm,width=8cm]{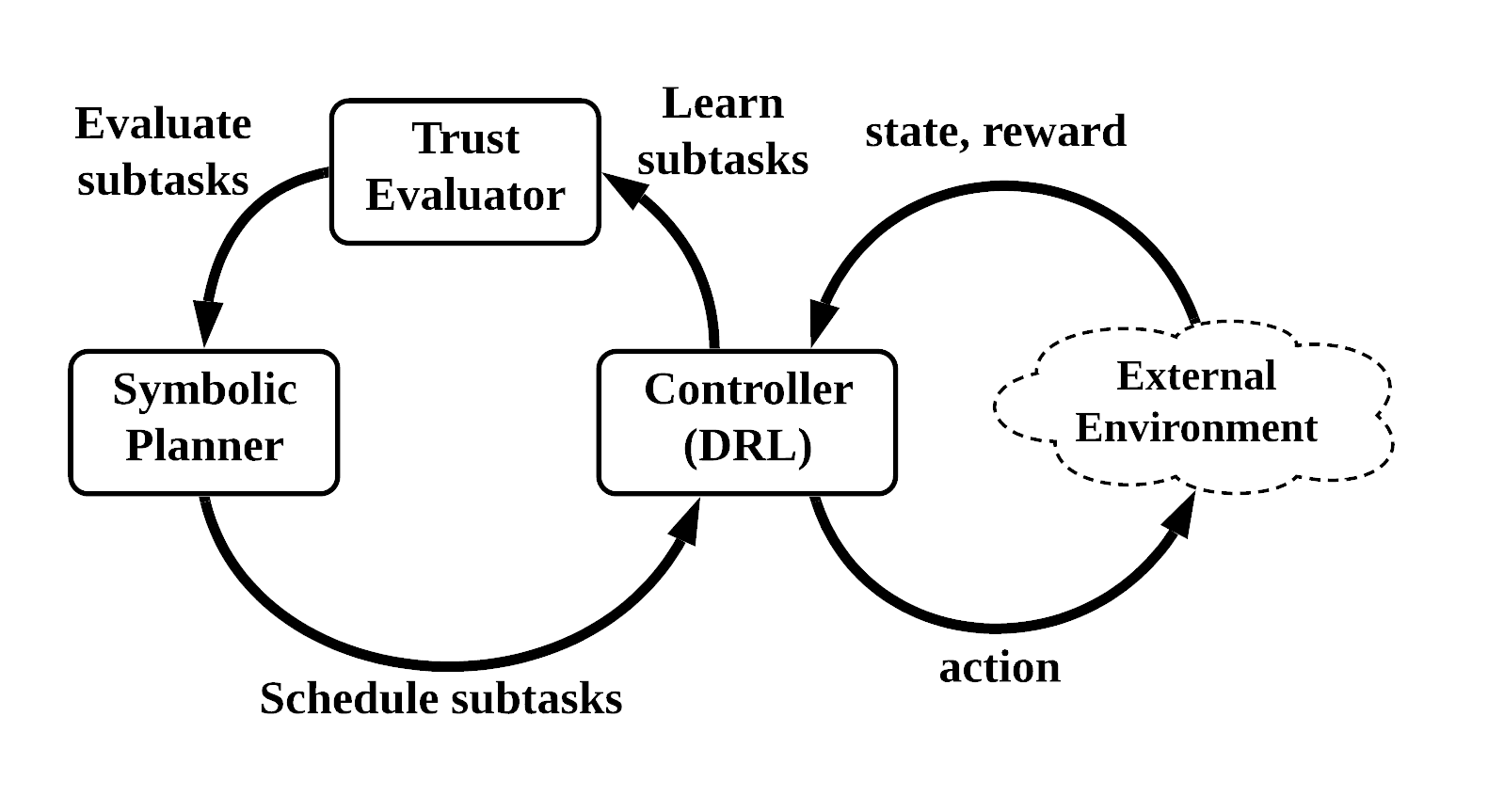}
\caption{Architecture illustration}
\label{fig:arch}
\end{figure}

With a symbolic representation given by the human expert, the TDM architecture is shown in Fig.~\ref{fig:arch}. A symbolic planner generates a high-level plan, i.e., a sequence of subtasks, based on the trust evaluator's evaluation feedback. 
We implement a mapping function to perform symbol grounding in our experiments, translating the problem domain to a symbolic representation. However, we assume such functions are generally available in different domains.
After symbol grounding, the subtasks in the plan will be sent to the controller to execute so that the sub-policies for the corresponding subtasks are learned. Since it is possible that the controller cannot successfully learn the sub-policies, a metric is introduced to evaluate the competence of learned sub-policies, such as the success ratio over several episodes. After all subtasks in the plan have been executed, the trust evaluator 
computes the trust score by utilizing the trustworthiness values for subtasks. The evaluation for subtasks is returned to the symbolic planner and is used to generate new plans by either exploring new subtasks or sequencing learned subtasks that supposedly can achieve a higher trust score in the next iteration.

\subsection{A Computational Framework for Trust Evaluation}

We formulate the trust evaluation with a success ratio of task execution, a simple measure of quality. Intuitively, it is easier to measure the success of interpretable tasks than non-interpretable tasks. By design, our plans and subtasks are interpretable, and therefore it is easier to devise the corresponding trust score in our problem.

Trust has been evaluated by using the internal utility function in our case to measure the plan's trust score. By doing so, trust evaluation can establish the connection between the trust metric and observation (trust evidence), and justify the decision made upon trust score. Therefore, the subtasks in a plan will be evaluated quantitatively, and the plan with a low trust score will have less ability to be considered for the final solution.

The internal utility score based on the success ratio alone is not a convincing measure of trust.
So we tie the trust evaluation to the interpretability perspective and promote the subtasks that are both successful and interpretable as trustworthy.
In essence, we try to solve an optimization problem to maximize the explicit trust score subject to the constraint of implicit interpretability:
\begin{equation*}
\max_{\Pi} \frac{1}{\vert \Pi \vert} \sum_{o \in \Pi} \epsilon(o) \text{ s.t. } I(\Pi) > \delta,
\end{equation*}
\noindent
where $\epsilon$ is the success ratio of a symbolic task $o$, $\Pi$ is a symbolic plan, $I$ is an oracle able to numerically qualify the interpretability of a symbolic plan, and $\delta$ is the interpretability threshold.

The success ratio with respect to interpretability is the key to our trust evaluation mechanism.
But interpretability is a qualitative measure, and it is difficult to formulate a traditional optimization approach to satisfy this constraint.
Instead, we use the symbolic representation to naturally induce interpretability in our problem.
We discuss the details of our approach in the subsequent sections.

\subsection{Interpretability Enhancement via Symbolic Representation}\label{sec:sr}

Let us consider a planning problem $(I, G, D)$.
Although a symbolic formulation is possible with various planning or action languages, we will use $\mathcal{BC}$ to represent $D$ as a demonstration.
Specifically, we add the following causal laws to $D$ to formulate gain rewards of executing actions and their effects on cumulative plan quality:
\begin{itemize}
\item For any symbolic state that contains atoms $\{A_1,\ldots, A_n\}$, $D$ contains static laws of the form:  
$$
 s~\iif~A_1,\ldots, A_n,~\hbox{for~state}~s\in\mathcal{S}.
$$
\item We introduce new fluent symbols of the form $\rho(s,a)$ to denote the gain reward at state $s$ following action $a$.
$D$ contains a static law stating that the gain reward is initialized optimistically by default to promote exploration and is denoted as~$\i{INF}$:
$$
\!\!\!\default~\rho(s,a)=\i{INF},~\hbox{for}~s\in\mathcal{S},a\in\sigma_A(D).
$$
\item We use fluent symbol $\i{quality}$ to denote the total trustworthiness value of a plan, termed as {\em plan quality} or {\em trust score}. $D$ contains dynamic laws of the form
$$
\!a~\causes~\i{quality}=C+Z~\iif~s,\rho(s,a)=Z,\i{quality}=C.\!
$$  
\item $D$ contains a set $P$ of facts of the form $\rho(s,a)=z$.
\end{itemize}
In our case, $I$ is still the initial symbolic planning state, but the goal state $G$ is also a linear constraint of the form 
\beq\i{quality} > \i{quality}(\Pi),
\eeq{goal}
for a symbolic plan $\Pi$ measured by the internal utility function $\i{quality}$ defined as
\beq
\i{quality}(\Pi) = \sum_{\langle s_{i},a_{i},s_{i+1}\rangle\in\Pi} \rho(s_{i},a_{i}).
\eeq{quality}
It should be noted that (\ref{goal}) in TDM doesn't have the logical constraint part and enables ``model-based exploration by planning'', which is more suitable for the problems where the trust score drives the agent's behavior.

With the help of declarative paradigms for modeling, symbolic representation has good interpretability in its nature as the hypotheses are understandable and interpretable.

\subsection{Implicit Interpretability Constraint Satisfaction: From Symbolic Transitions to Subtask Options}

\begin{figure}[!t]
	\centering
	\includegraphics[height=4cm,width=8cm]{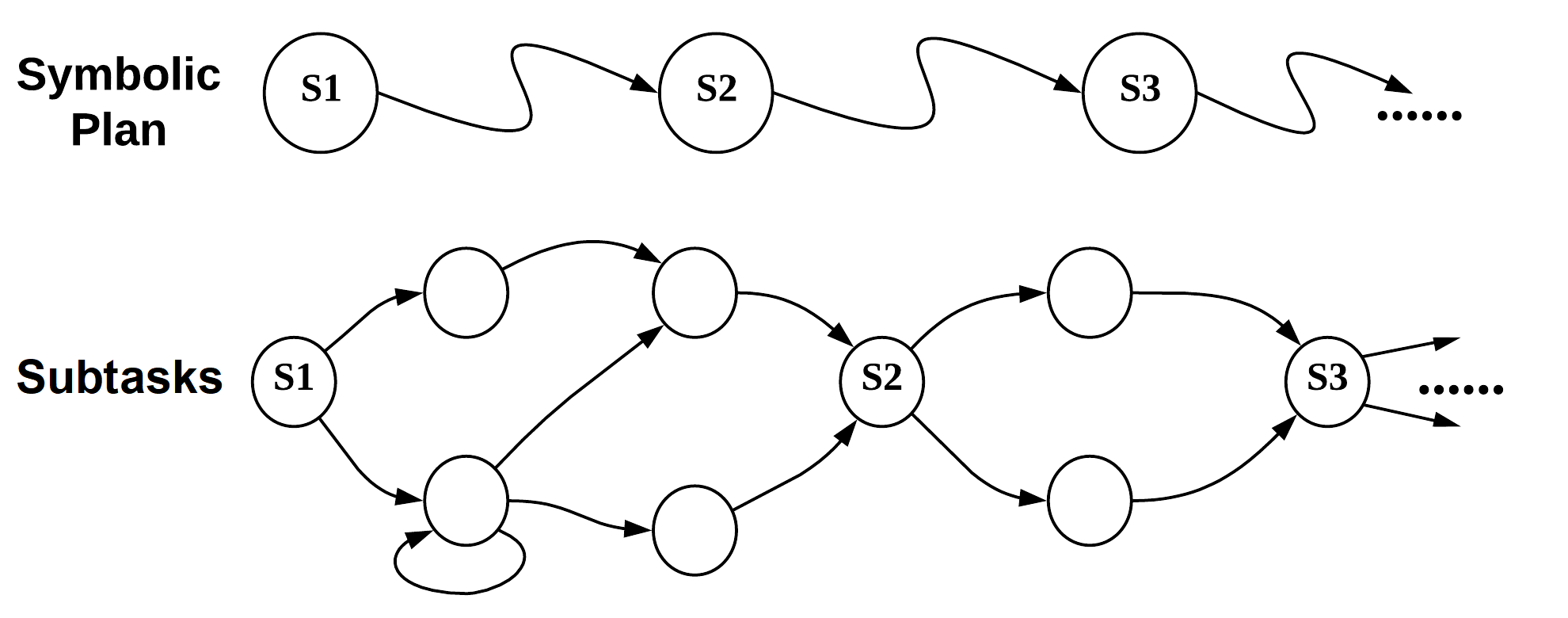}
	\caption{The mapping from a symbolic transition path to subtasks}
	\label{fig:option}
\end{figure}

Our discussion of trust evaluation and interpretability enhancement has been restricted to general machine learning.
Symbolic planning can be used in many problem domains, e.g., image classification, and success ratio may be replaced with a suitable metric, such as accuracy, in the respective domain.
Here, we narrow our focus to a sequential decision-making problem, which will allow us to devise a specific algorithm based on the TDM framework.

Let us define a symbolic mapping function as $\mathbb{F}:\mathcal{S}\times\widetilde{\mathcal{S}}\mapsto\{\t,\f\}$.
If a symbolic state $s \in \mathcal{S}$ corresponds to $\tilde{s} \in \mathcal{\tilde{S}}$, then $\mathbb{F} = \t$. $\mathbb{F} = \f$ if otherwise.
We assume an oracle exists to determine whether the symbolic properties specified as fluent atoms of the form $f=v$ in $s$ are true in $\tilde{s}$.
Due to advances in computer vision, an oracle such as a perception module for object recognition in images is not uncommon.

An MDP can be considered as a flat decision-making system where the decision is made at each time step. On the contrary, humans make decisions by incorporating temporal abstractions.
A {\em subtask policy}~$(I,\pi ,\beta)$ is temporally extended course of action where a policy $\pi: {\mathcal{S}} \times {\mathcal{ A }} \mapsto {{ [0, 1]}}$, a termination condition~$\beta: {\mathcal{S}} \mapsto {{ [0, 1]}}$, and an initiation set ${I} \subseteq {\mathcal{S}}$.
A subtask policy~$(I,\pi ,\beta)$ is available in state $s_t$ if and only if ${s_t} \in I$.
When a subtask policy is initiated, the subtask is executed through actions stochastically selected according to~$\pi$ until a terminating symbolic state is reached according to $\beta$.

Given $\mathbb{F}$ and a pair of symbolic states $s,s'\in\mathcal{S}$, we can induce a semi-Markov subtask policy, as a triple $(I,\pi,\beta)$ where the initiation set $I=\{\tilde{s}\in\widetilde{\mathcal{S}}:\mathbb{F}(s,\tilde{s})=\t\}$, $\pi:\widetilde{\mathcal{S}}\mapsto\widetilde{\mathcal{A}}$ is the intra-subtask policy, and $\beta$ is the termination condition such that 
$$\beta(\tilde{s'})=\left\lbrace
\ba{ll}
1 & \mathbb{F}(s',\tilde{s'})=\t, \hbox{for}~\tilde{s'}\in\widetilde{\mathcal{S}}\\
0 &\hbox{otherwise}
\ea\right.
$$  
The formulation above maps symbolic transition to a similar structure of subtask policies.
It implies that the execution of the subtask will be feasible if the termination condition is $1$, while it will be infeasible if the termination condition is $0$.
Therefore, the interpretability of the subtasks depends on how they contribute to the plan's symbolic transitions.
The interpretability of the feasible subtasks is guaranteed by the interpretability of the corresponding symbolic transition. In contrast, the infeasible subtasks are abandoned as un-interpretable as they do not contribute to any interpretable symbolic plan. 

By mapping the tasks of the symbolic plan to the subtask policies, we incorporate the hierarchical structure to the decision-making where the symbolic task controls the higher-level planning, and the corresponding subtask policy controls the primitive actions required to accomplish the task.
As illustrated in Fig.~\ref{fig:option}, the subtask policies exhibit Markovian property at the symbolic task level and thus executed sequentially according to the plan.

\subsection{Explicit Success Ratio Maximization}

We first assign $t_e(\tilde{s'})$ with some trustworthiness values in a binary form, which is used for computing the trust score.
It is defined as: 
\beq
t_e(\tilde{s'})=\left\lbrace
\ba{ll}
+1, & \beta(\tilde{s'})=1\\
-1, & \hbox{otherwise}
\ea\right.
\eeq{intrinsic}
which means that the trustworthiness value will be $+1$ if the subtask terminated at $\tilde{s'}$ can be achieved, otherwise it will be $-1$. {\color{black}{Other more sophisticated distribution (e.g., Bernoulli distribution) can also be modeled other than this 0-1 distribution model in practice.}}

We further define $r_e(s,o)$ as
$
r_e(s,o) = f(\epsilon),
$
where $f$ is a function of $\epsilon$, and $\epsilon$ is the success ratio that denotes the average rate of completing the subtask successfully over the previous 100 episodes.
This metric is used to measure whether the subtask $o$ at symbolic state $s$ is learnable or not. This is different from the definition of $t_e(\tilde{s'})$ since $r_e(s,o)$ requires reliably achieving the subtask after the training by episodes and keeping the success ratio above a certain threshold. Therefore, $f$ is defined as
\beq
f(\epsilon)=\left\lbrace
\ba{ll}
-\psi & \epsilon<\Epsilon\\
r(s,o) & \epsilon\ge \Epsilon
\ea\right.
\eeq{extrinsic}
where $\psi$ is a large positive numerical value, $r(s,o)$ is the reward obtained from the MDP environment by following the subtask $o$, and the hyper-parameter$\Epsilon$ is a predefined threshold of success ratio. In our experiments, we set $\Epsilon = 0.9$ according to empirical performance observations. Intuitively, it means if the sub-policy can reliably achieve the subtask after the training by episodes, then the trustworthiness value of $r_e(s,o)$ at $s'$ reflects true cumulative reward from the MDP environment by following the subtask; otherwise, the trustworthiness value will be very low (as a large negative number), indicating that the sub-policy performs badly and is probably not learnable. A plan $\Pi$ of $(I,G,D)$ is considered to be {\em optimal} iff
$\sum_{\langle s,a,s'\rangle} r_e(s, o)$ is maximal among all plans.

\section{Algorithm Design}\label{sec:alg}

We present a learning algorithm for the TDM framework in Algorithm~\ref{algexec} and show TDM's planning and learning process.
Each episode starts with the symbolic planner generating a symbolic plan $\Pi_t$ for the problem $(I, G, D)$ (Line 4).
The symbolic transitions of $\Pi_t$ are mapped to subtasks (Line 10) to be learned by a controller (Line 11).
The controller performs deep Q-learning with $t_e$ using experience replay (Lines 12--15) to estimate 
the $Q$ value $Q(\tilde{s},\tilde{a};o)\approx Q(\tilde{s},\tilde{a};\theta,o)$, where $\theta$ is the parameter of the non-linear function approximator.
The observed transition $(\tilde{s}_t,\tilde{a}_t,r_e(\tilde{s}_{t+1},g),\tilde{s}_{t+1})$ is stored as an experience in $\mathcal{D}_o$.
The loss at $i^\text{th}$ iteration is calculated as an expectation over the collected experience and is given by

\beq
\ba{l}
L(\theta;o) = \hbox{E}_{(\tilde{s},\tilde{a},g,t_e,\tilde{s}')\sim D_o}[r_e + \gamma\max_{\tilde{a'}} Q(\tilde{s},\tilde{a'};\theta_{i-1},o) \\
\text{}-Q(\tilde{s},\tilde{a};\theta_i,o)]^2.
\ea
\eeq{dqn}

When the inner loop terminates, a symbolic transition $\langle s_t,a_t,s_{t+1}\rangle$ for the subtask $o_t$ is completed, and we are ready to update the trust score of the subtask (Line 18).
Although various learning methods, such as Q-learning, may be used here, we use R-learning \cite{sm:mlj96} to evaluate the trust score as an average reward case rather than a discounted reward case. With R-learning, the update rule becomes
\beq
\ba{rl}
R_{t+1}(s_{t},o_{t})\!\!\!&\!\!\xleftarrow{\alpha} r_e - \rho_t^{o_{t}}(s_{t})+\max_{o} R(s_t,o)\\
\rho_{t+1}^{o_{t}}(s_{t})\!\!\!&\!\!\xleftarrow{\beta}r_e+\max_{o} R_t(s_{t+1},o) - \max_{o} R_t(s_{t},o)
\ea
\eeq{riter1}
The {\em quality} of $\Pi_t$ is measured by (\ref{quality}) (Line 20) and the trust score of the plan is updated according to $\i{quality}(\Pi_t)$ (Line 21).
The symbolic formulation is then updated with the learned ~$\rho$ values (Line 22).
With the symbolic formulation changed, an improved plan may be generated in the next episode.
The loop terminates when no such improvement can be made.
\begin{figure}[t!]
\removelatexerror
\begin{algorithm}[H]
{\small
  \caption{TDM Planning and Learning Loop}
  \label{algexec}
  \begin{algorithmic} 
    \renewcommand{\algorithmicrequire}{\textbf{Input:}}
    \renewcommand{\algorithmicensure}{\textbf{Output:}}
    \REQUIRE $(I,G,D,\mathbb{F})$ where $G=(\i{quality}>0)$, and an exploration probability $\epsilon$
    \ENSURE An optimal symbolic plan $\Pi^*$
    \STATE $P_0\Leftarrow \emptyset$, $\Pi\Leftarrow \emptyset$
    \WHILE{True}
      \STATE $\Pi^*\Leftarrow \Pi$
      \STATE Take $\epsilon$ probability to solve planning problem and obtain a plan $\Pi\Leftarrow\textsc{clingo}.\i{solve}(I,G,D\cup P_t)$
      \IF {$\Pi=\emptyset$}
          \RETURN $\Pi^*$
      \ENDIF
      \FOR {symbolic transition $\langle s, a, s'\rangle\in\Pi$}
      \STATE Obtain current state $\tilde{s}$
      \STATE Correspond to subtask $o$ by using $\mathbb{F}$ to obtain initiation set and terminate condition
      \WHILE {$\beta(\tilde{s})\neq 1$ and maximal step is not reached}
      \STATE Pick up an action $\tilde{a}$ and obtain transition $(\tilde{s},\tilde{a},\tilde{s'},t_e(\tilde{s'}))$
      \STATE Store transition in experience replay buffer $\mathcal{D}_o$
      \STATE Estimate $Q(\tilde{s},\tilde{a};\theta,o)$ by minimizing loss function~(\ref{dqn}) when there are sufficient samples in $\mathcal{D}_o$
      \STATE Update current state $\tilde{s}\Leftarrow\tilde{s'}$
      \ENDWHILE
      \STATE Calculate $r_e(s,o)$ with trustworthiness values
      \STATE Update $R(s,o)$ and $\rho^o(s)$ using (\ref{riter1}).
      \ENDFOR
      \STATE Calculate quality of $\Pi$ by (\ref{quality}).
      \STATE Update planning goal $G\Leftarrow (\i{quality}> \i{quality}_t(\Pi))$.
      \STATE Update facts $P_t\Leftarrow \{\rho(s,a)=z:\langle s,a,s'\rangle\in\Pi, \rho_t^{a}(s)=z\}$
    \ENDWHILE
  \end{algorithmic}}
\end{algorithm}
\end{figure}
The algorithm guarantees symbolic level optimality conditioned on R-learning convergence. 

\noindent
\begin{theorem}[Termination]
If the trust evaluator's R-learning converges, Algorithm~\ref{algexec} terminates iff an optimal symbolic plan exists.
\end{theorem}

\noindent{\textbf{Proof.}} When R-learning converges, for any transition $\langle s,a,t\rangle$, the increment terms in (\ref{riter1}) diminish to $0$, which implies 
\beq
\!\!\!
\ba{c}
R(s,o) = \max_{o'} R(s,o'),\\
\rho^a(s) = r_e(s,o) - \max_{o'} R(s,o') + \max_{a'} R(t,o')
\ea
\eeq{op1}
Algorithm~\ref{algexec} terminates iff there exists an upper bound of plan quality iff there does not exist a plan with a loop $L$ such that
$
\sum_{\langle s,a,t\rangle\in L} \rho^a(s)> 0.
$
By (\ref{op1}), it is equivalent to
$
\sum_{\langle s,o,t\rangle\in L} (r_e(s,o)-R(s,o)+ \\ R(t,o)) \le 0$ iff
$\sum_{\langle s,o,t\rangle\in L} r_e(s,o) - R(s_{|L|},o) + R(s_0,o) \le 0.
$
Since $L$ is a loop, $s_{|L|} = s_0$, so 
$
\sum_{\langle s,a,t\rangle\in L} r_e(s,o) \\ \le 0
$
iff any plan $\Pi$ does not have a positive loop of cumulative reward iff optimal plan exists, which completes the proof.

\begin{theorem}[Optimality]
If the trust evaluator's R-learning converges, when Algorithm~\ref{algexec} terminates, $\Pi^*$ is an optimal symbolic plan.
\end{theorem}

\noindent{\textbf{Proof.}}~By \citeauthor{lee13}~\cite[Theorem 2]{lee13}, $\Pi^*$ is a plan for planning problem $(I, G, D)$.
For $\Pi_o$ returned by Algorithm~\ref{algexec} when it terminates,
$\i{quality}(\Pi) \le \i{quality}(\Pi^*)~\hbox{for~any}~\Pi$ iff $$\sum_{\langle s,a,t\rangle\in\Pi}\rho^a(s) \le \sum_{\langle s,a,t\rangle\in\Pi_o}\rho^a(s).$$ By (\ref{op1}), the inequality is equivalent to
$$
\sum_{\langle s,a,t\rangle\in\Pi}r_e(s,o) + R(s_{|\Pi|},o) \le \sum_{\langle s,a,t\rangle\in\Pi_o}r_e(s,o) + R(s_{|\Pi^*|},o).
$$
Since $s_{|\Pi|}$ and $s_{|\Pi^*|}$ are terminal states of each symbolic plan with no subtask policies available, we have
$
\sum_{\langle s,a,t\rangle\in\Pi}r_e(s,o)  \le \sum_{\langle s,a,t\rangle\in\Pi_o}r_e(s,o).
$
This completes the proof.

\section{Experiment}\label{sec:exp}

We use the Taxi domain \cite{barto-sm:hrl} to demonstrate the behavior of learning and planning based on the trust score, Grid World \cite{leonetti2016synthesis} to show the ability to learn unmodeled domain knowledge, and on Montezuma's Revenge~\cite{dqn:nature:2015} for interpretability and data efficiency. We use 1M to denote $1$ million and 1k to denote $1000$. 

\subsection{Taxi Domain} 

\begin{figure}[!t]
\centering
\begin{subfigure}{.30\textwidth}
  \centering
  \includegraphics[width=4.5cm, height=4.2cm]{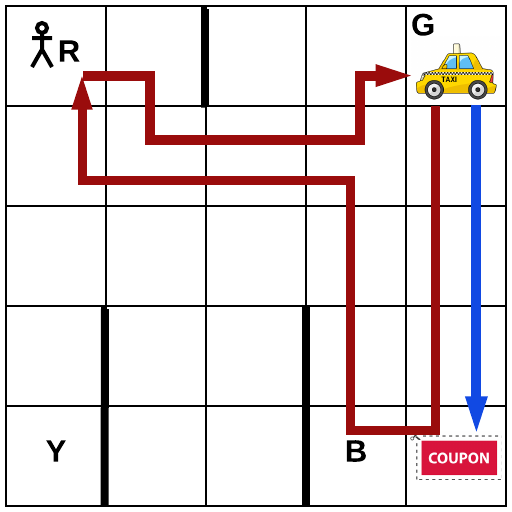}
  \caption{Taxi Domain}
  \label{fig:t2}
\end{subfigure}
\quad
\begin{subfigure}{.50\textwidth}
  \centering
  \includegraphics[width=7.2cm,height=5.5cm]{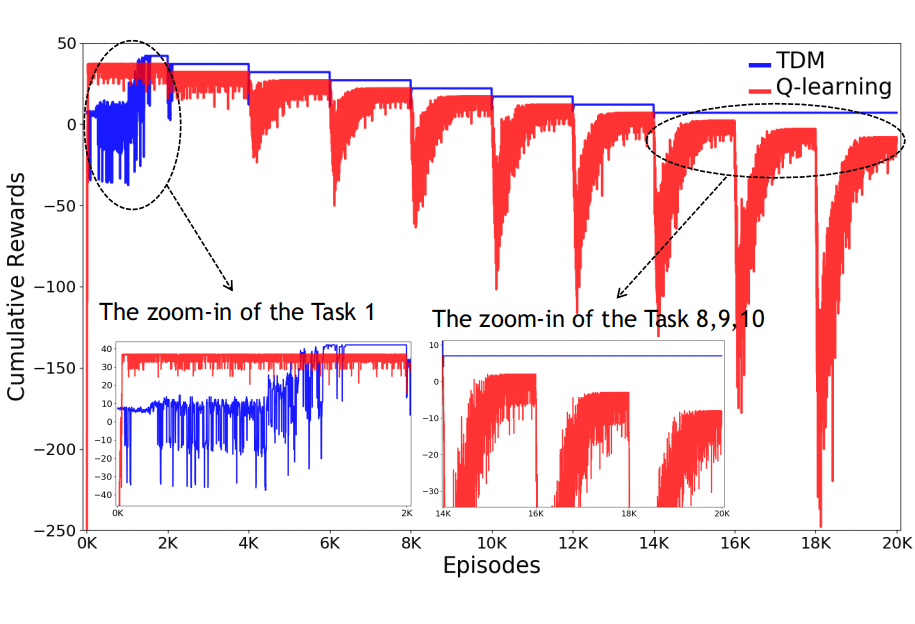}
  \caption{Learning Curve}
  \label{fig:taxi-reward}
\end{subfigure}
\caption{Experimental Results on Taxi Domain}
\label{fig:taxi}
\end{figure}

The objective of the Taxi domain is to maximize reward by successfully picking-up and dropping-off a passenger at a specified destination on a grid map (Fig.~\ref{fig:t2}).
A taxi agent moves from a cell to an adjacent cell with -1 reward collected at each time step.
The agent receives a reward of 50 for dropping the passenger off at the destination while receiving a reward of -10 when attempting to pick-up or drop-off at an incorrect location.
Additionally, there is a one-time reward coupon the taxi can collect at $(4,4)$. With a reward of 10, we will observe the behavior of learning agents when the environment changes.

\noindent\textbf{Setup.}
For each episode, the taxi starts at $(0,4)$, and we will call every 2000 episodes a task.
The experiment consists of 10 tasks where the reward for the successful drop-off decreases in each task by 5, i.e., 50 in Task 1, 45 in Task 2, and so forth.
Other reward settings stay constant throughout the tasks.
We compare TDM with Q-learning.

\noindent\textbf{Experimental Results.} 
First, let us consider the optimal policies as shown in Fig.~\ref{fig:t2}.
The dark red colored route that collects the coupon, picking up, and then dropping off the passenger is the optimal policy from Task 1 to Task 7.
From Task 8 and onward, the reward for the passenger drop-off ($\leq 15$) is no longer worth the effort, and hence, the optimal policy changes to simply collecting the coupon as indicated by the blue line.
As shown in the learning curve in Fig.~\ref{fig:taxi-reward}, averaged over 10 runs, TDM successfully learns the optimal policy for all Tasks. As noted, the optimal policy changes at Task 8, and TDM quickly abandons the sub-tasks that lead to a sub-optimal policy and learns the new optimal policy.
All these are in contrast to Q-learning. While Q-learning converges faster than TDM (see the zoom-in of Task 1 in Fig.~\ref{fig:taxi-reward}) because its model-free exploration is more aggressive than TDM's exploration guided by the trust-score-based planning, Q-learning often fails to learn the optimal policy.
Moreover, 
once Q-learning converges to a policy, it is unable to learn a new one when the environment changes.
On the other hand,
the trust evaluation mechanism in TDM allows the agent to be flexible and abandon the plan with lower trust score.
In practice, the reward is often collected from interaction and changes over time in a dynamic environment. This experiment shows TDM may adapt well in such situations.

\subsection{Grid World}

\begin{figure*}[!t]
\centering
\begin{subfigure}{.45\textwidth}
  \centering
  \includegraphics[width=4.5cm, height=4.5cm]{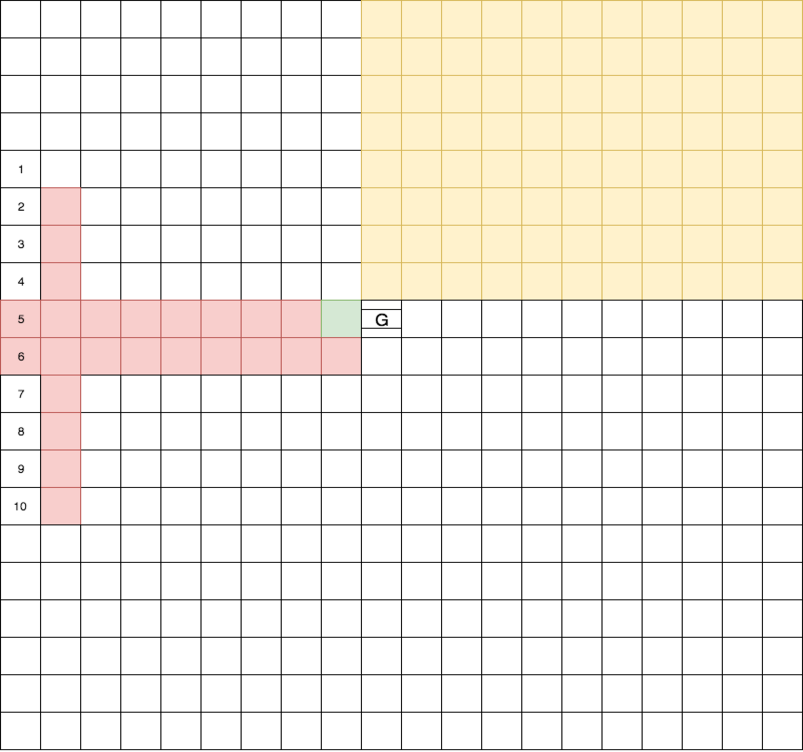}
  \caption{Grid World}
  \label{fig:gridworld-exp}
\end{subfigure}
\quad
\begin{subfigure}{.45\textwidth}
  \centering
  \includegraphics[width=4.5cm, height=4.5cm]{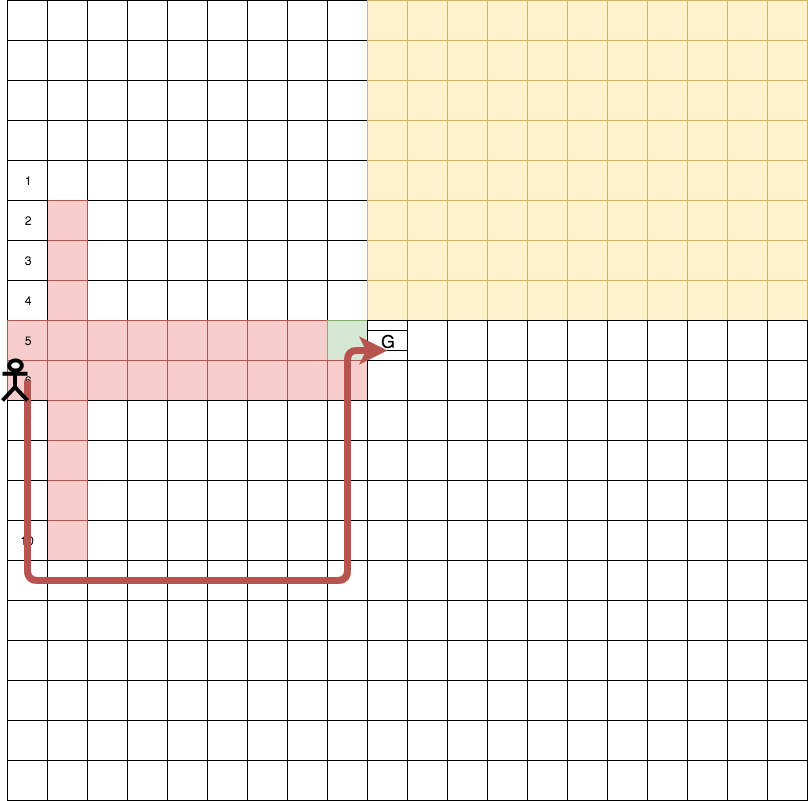}
  \caption{A solution}
  \label{fig:gridworld-exp-route}
\end{subfigure}
\begin{subfigure}{.45\textwidth}
  \centering
  \includegraphics[width=6.8cm, height=4.5cm]{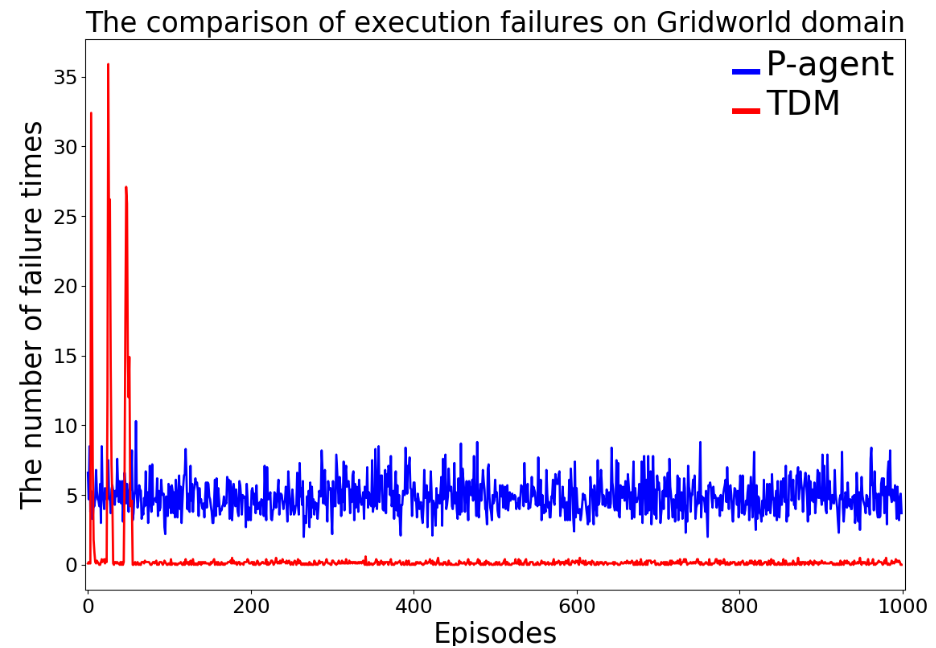}
  \caption{{\small Execution failure of TDM and Planning}}
  \label{fig:gridworld-failures}
\end{subfigure}
\quad
\begin{subfigure}{.45\textwidth}
  \centering
  \includegraphics[width=6.2cm, height=4.5cm]{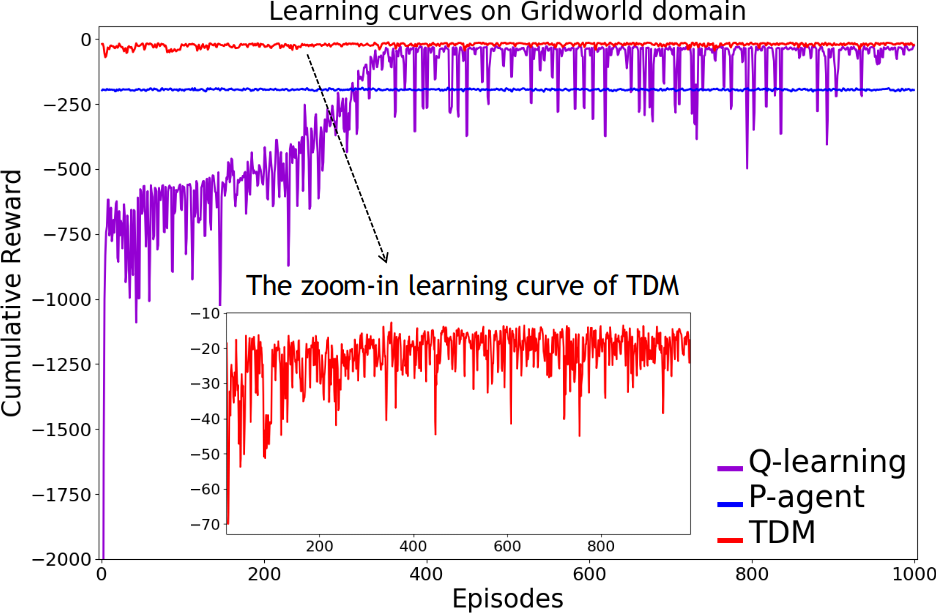}
  \caption{{\small Learning Curve}}
  \label{fig:gridworld-reward}
\end{subfigure}
\caption{Grid World}
\end{figure*}

TDM can learn domain details that are not modeled into symbolic knowledge. We demonstrate this behavior with the Grid World adapted from 
\citeauthor{leonetti2016synthesis}~\cite{leonetti2016synthesis}, shown in Fig.~\ref{fig:gridworld-exp}. An agent must navigate to (9,10), which can only be entered through a door at (9, 9). At the door, the agent must grab a doorknob, turn it, and then push the door to reach the goal, all of which may fail and incur a -10 penalty. As in the Taxi domain, every movement has a reward of -1.

\noindent\textbf{Setup.} Horizontal and vertical bumpers represent the domain details that the agent needs to learn. The agent receives an additional penalty for a movement into a bumper. Penalties are -30 and -15 for red and yellow bumpers, respectively. The agent starts at random on one of the numbered grids in the first column. We use Q-learning and a standard planning agent (P-agent) as baselines for this domain. P-agent generates plans using \textsc{Clingo} and executes the plans without any learning capability.

\noindent\textbf{Experimental Results.} Learning curves of cumulative reward are shown in Fig.~\ref{fig:gridworld-reward}. TDM achieves the optimal behavior: it avoids the bumpers and learns to reliably open and push through the door (e.g., Fig.~\ref{fig:gridworld-exp-route}), surpassing Q-learning in both the rate of learning and variance of the cumulative reward. P-agent prefers shortest plans which are not ideal plans in this case. P-agent relies on re-planning upon encountering failures without learning. As such, the number of execution failures to enter the door remains high throughout episodes, as shown in Fig.~\ref{fig:gridworld-failures}. On the other hand, TDM abandons unsuccessful plans due to their low trust scores and quickly learns to correctly open the door and reach the goal.

\subsection{Montezuma's Revenge}

In ``Montezuma’s Revenge,'' the player controls a game character through a labyrinth avoiding dangerous enemies and collecting items that are helpful to the player. We focus on the first room of the labyrinth. Here, the player has to collect a key in the room to unlock the door to the next room. In a typical setting for DRL, an agent receives +100 reward for collecting the key and +300 for opening the door with the key. This is a challenging domain for DRL as it involves a long sequence of high-level tasks and many primitive actions to achieve those tasks.
{
This is a challenging domain where the vanilla DQN often fails to learn~\cite{dqn:nature:2015}.
}

\begin{table}[!t]
{\scriptsize
\begin{center}
\begin{tabular}{ c|c| c }
\hline\hline No.& Layer &  Details  \\
\hline
1& Convolutional & 32 filters, kernel size=8, stride=4, activation='relu' \\
2& Convolutional & 64 filters, kernel size=4, stride=2, activation='relu' \\
3& Convolutional & 64 filters, kernel size=3, stride=1, activation='relu' \\
4&  Fully Connected & 512 nodes, activation='relu'\\
5&  Output & activation='linear' \\
\hline\hline
\end{tabular}
\end{center}}
\caption{Neural Network Architecture for Montezuma's Revenge}
\label{tab:montezuma-nn}
\end{table}

\noindent{\bf Setup.} Our experiment setup follows the DQN controller architecture~\cite{kulkarni2016deep} with double-Q learning \cite{van2016deep} and prioritized experience replay \cite{schaul2015prioritized}. The architecture of the deep neural networks is shown in Table~\ref{tab:montezuma-nn}.
The experiment is conducted using Arcade Learning Environment (ALE)~\cite{bellemare2013arcade}.
We have implemented the symbolic mapping function $\mathbb{F}$ based on ALE API.
The binary trustworthiness value follows (\ref{intrinsic}) for when a subtask successfully completes or the agent loses its life.
The subtask trust score follows (\ref{extrinsic}) where $\psi=100$ and define $r(s,o)=-10$ for $\epsilon>0.9$ to encourage shorter plan.
We use hierarchical DQN (hDQN) \cite{kulkarni2016deep} as the baseline.

\noindent{\bf Symbolic Representation.} 
We represent the transition dynamics and domain knowledge in action language $\mathcal{BC}$. There are $6$ pre-defined locations or objects: middle platform ({\tt mp}), right door ({\tt rd}), left of rotating skull ({\tt ls}), lower left ladder ({\tt lll}), lower right ladder ({\tt lrl}), and key ({\tt key}). Note that the number of predefined locations or objects depends on the users and their domain knowledge.
We then formulate 13 subtasks based on the symbolic locations. The corresponding symbolic transitions from the mapping function $\mathbb{F}$ are shown in Table~\ref{tab:montezuma-subgoal}.
The difference between our and hDQN's subtask definitions should be noted. A subtask is only associated with an object in hDQN; However, in our work, it is defined as a symbolic transition with an initiation set and termination condition based on the states satisfying symbolic properties.
With this informative symbolic representation, our approach is more general, descriptive, and interpretable. It also makes sub-policy for each subtask to be more easily learned and subtasks more easily sequenced.

\begin{table}[!t]
{\scriptsize
\begin{center}
\begin{tabular}{ c|c| c| c }
\hline\hline No.& Subtask &  Policy learned & In the optimal plan  \\
  \hline
1&  MP to LRL, no key & \checkmark & \checkmark \\
2&  LRL to LLL, no key & \checkmark & \checkmark\\
3&  LLL to key, no key & \checkmark &\checkmark\\
4&  key to LLL, with key& \checkmark & \checkmark \\
5&  LLL to LRL, with or without key & \checkmark & \checkmark\\
6&  LRL to MP, with or without key & \checkmark & \checkmark \\
7&  MP to RD, with key & \checkmark & \checkmark\\
\hline
8&  LRL to LS, with or without key & \checkmark &  \\
9&  LS to key, with or without key & \checkmark &  \\
10& MP to RD, no key & \checkmark &  \\
\hline
11& LRL to key, with or without key &  &  \\
12 & key to LRL, with key &  &   \\
13 & LRL to RD, with key &  &  \\
\hline\hline
\end{tabular}
\end{center}}
\caption{Subtasks for Montezuma's Revenge}
 \label{tab:montezuma-subgoal}
\end{table}

\begin{figure*}[!t]
\centering
\begin{subfigure}{.45\textwidth}
 \centering
 \includegraphics[width=6.5cm, height=4.8cm]{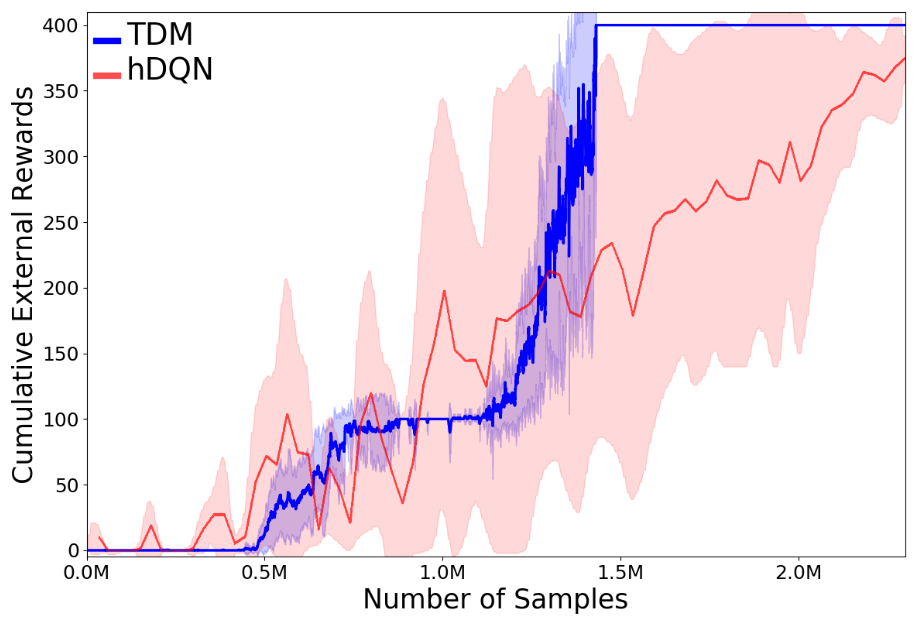}
   \caption{Learning Curve}
   \label{fig:cumulative}
\end{subfigure}
\quad
\begin{subfigure}{.45\textwidth}
 \centering
  \includegraphics[width=7cm, height=4.8cm]{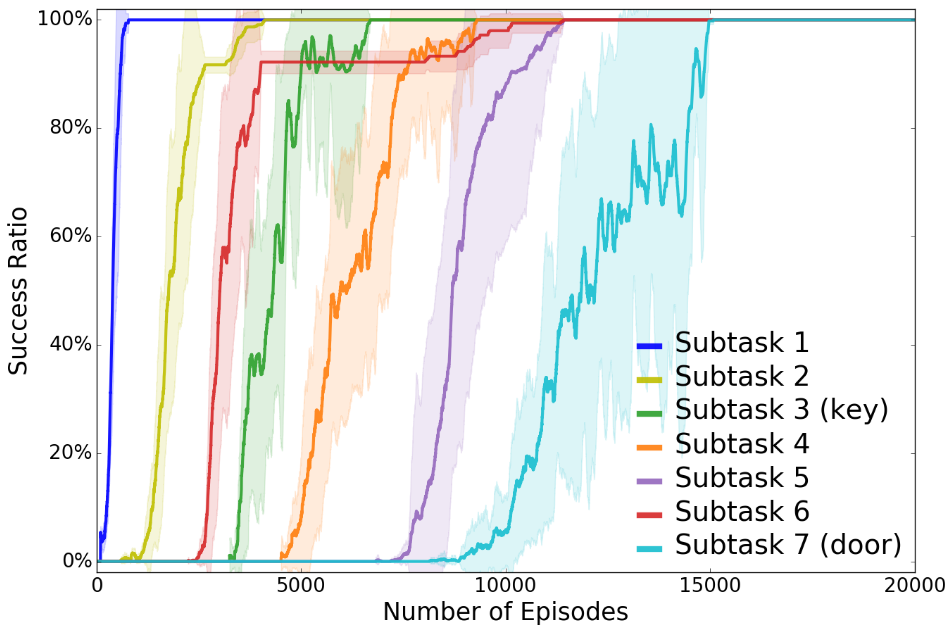}
  \caption{Success Ratio}
  \label{fig:success}
\end{subfigure}
 \begin{subfigure}{.45\textwidth}
 \centering
\includegraphics[width=6.8cm, height=4.8cm]{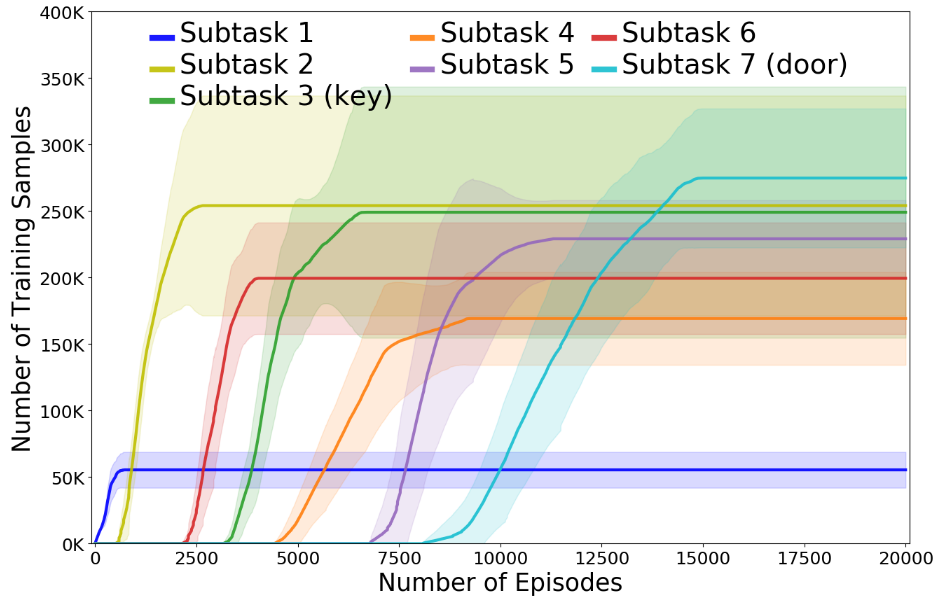}
   \caption{Samples Required}
   \label{fig:sample}
 \end{subfigure}
\quad
 \begin{subfigure}{.45\textwidth}
 \centering
  \includegraphics[width=6.5cm, height=4.5cm]{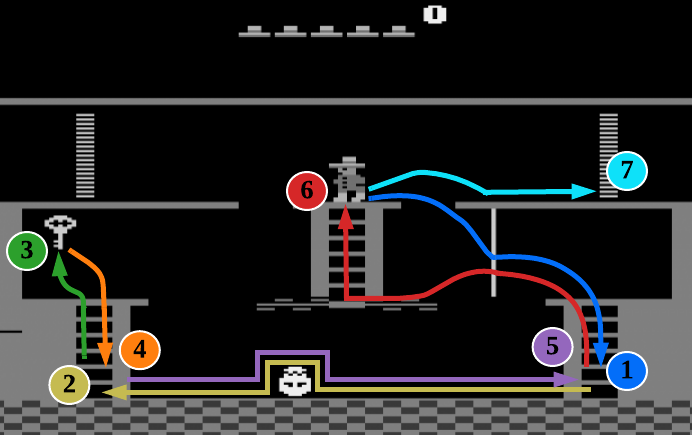}
   \caption{Final Solution}
   \label{fig:final}
\end{subfigure}
\caption{Experimental Results on Montezuma's Revenge}
\end{figure*}

\noindent\textbf{Experimental Results.} 
It is easy to see that TDM is more data-efficient than the baseline hDQN from the learning curve (Fig.~\ref{fig:cumulative}).
Interpretability requires a qualitative analysis. We will emphasize how trust score guides TDM on planning, learning, and sequencing of the subtasks, and therefore learns the optimal behavior, all of which are visualized in our figures.  

The results have been averaged over 10 runs and we shown mean-variance plots in the Fig.~\ref{fig:cumulative},~\ref{fig:success},~\ref{fig:sample}.
The description of subtasks can refer to both Fig.~\ref{fig:final} and Table~\ref{tab:montezuma-subgoal}.
The environment rewards are only given for completing Subtask 3 (picking up the key, $+100$) and Subtask 7 (opening the right door, $+300$). Since other subtasks do not receive any reward from the environment, we can infer from Fig.~\ref{fig:cumulative} that TDM first learns the plan to sequence Subtasks 1--3. The agent learns other subtasks through exploration as the trust-score-based planning encourages executing untried subtasks and try different locations until it reaches the door. At that point, we can see in the learning curve that the agent quickly converges to the optimal plan that sequences through Subtasks 1--7 (Fig.~\ref{fig:final}), resulting in the maximal reward of $400$. In contrast, hDQN does not reliably achieve the maximal reward even after TDM stably converges. Also, it's difficult to see how hDQN sequenced through its subtasks due to high variance. TDM achieves a smaller variance than hDQN partially because our definition of the subtask is easier to learn than the one defined in hDQN, leading to more robust and stable learning.

The symbolic planning of TDM allows flexibility in reusing the learned subtasks in various plans. Fig.~\ref{fig:success} shows that TDM learns Subtask 6 before Subtask 3, but in the optimal plan, Subtask 3 must be sequenced before Subtask 6. This means Subtask 6 has been learned as a part of a different plan but re-selected in the optimal plan, which is possible because subgoals of the subtasks are associated by their starting states and only activated by the planner when the agent satisfies the starting states.

During the experiment, Subtasks 1--10 are successfully learned by controllers (or DQNs), with 7 of them being selected in the final solution with achieving a success ratio of $100\%$, shown in Fig.~\ref{fig:success}. 
TDM prunes other Subtasks 8--13 based on the low trust score achieved during training.
Subtask 8, from the lower right ladder to the left of the rotating skull, reaches a success ratio of $0.9$ but later quickly drops back to $0$, due to the instability of DQN. 
Subtasks 9 and 10 reach the required success ratio but are discarded as they do not contribute to the optimal plan, whereas the success ratio on subtasks 11--13 are poor, suggesting they are difficult to learn.

\noindent\textbf{Interpretability and Trust}
In our proposed framework, the evaluation criteria are designed through human involvement, based on the human expert's interpretation of the agent's execution. The symbolic representation can then be refined to improve the agent's performance. Therefore, the agent's decision-making's trustworthiness is rooted in the trust we place in the underlying symbolic representation.
Although many existing interpretable frameworks focus on simplifying the underlying model to generate human-understandable explanations 
\cite{ribeiro2016should, lipton2016mythos, wood2018transparent}, 
we show that the trade-off between the interpretability and performance of the model is not always necessary. In our case, a performance-based trust metric is used to guide the learning agent to maximize its performance and justify the subtasks it learns. This is, in essence, a separation of descriptive and persuasive explanation tasks \cite{herman2017promise}, where we handle the descriptive task through symbolic representation and perform persuasive explanation through the use of the trust score..

We now relate these concepts to the comparison of our framework to the baseline. In both hDQN and TDM, the bounding box defined by a human expert is utilized, considering the locations of objects only. As a result, the bounding box can be meaningful and interpretable with the human expert's domain knowledge. The subtasks of hDQN are only associated with the bounding box. However, TDM utilizes symbolic representation to derive subtasks except for the bounding box and can provide a relational perspective, such as objects/entities and their relations. This makes the subtasks of TDM more descriptive since a symbolic state may contain more rich information than hDQN. According to the trust evaluation on subtasks, trust scores would motivate the TDM agent to maximize its performance by executing the learnable subtasks while discarding the unlearnable ones. The experiment results show TDM's persuasive explanation is more convincing than hDQN. Also, the variance about the curves (Fig.~\ref{fig:cumulative}) demonstrates the degree of robustness when the agent faces uncertainties. The summarization is shown in Table~\ref{tab:compare-tdm-hdqn}.

\begin{table}[!t]
{\scriptsize
\begin{center}
\begin{tabular}{|c|c|c|c|}
\multicolumn{2}{c}{} & \multicolumn{1}{c}{TDM} & \multicolumn{1}{c}{hDQN} \\
\hline
\multirow{3}{*}{\textbf{Interpretability}}&  Interpretable & Yes & Yes \\
\cline{2-4}
& Descriptive & Strong & Weak \\
\cline{2-4}
& Persuasive & Strong & Weak\\
\hline
\multirow{1}{*}{\textbf{Trustworthiness}} & Robustness & Strong & Weak \\
\hline
\end{tabular}
\end{center}}
\caption{Comparison of TDM and hDQN}
 \label{tab:compare-tdm-hdqn}
\end{table}

\section{Conclusions}\label{sec:conclusion}
Since interpretability of the high-level behavior is critical to enhancing trust in a hierarchical sequential decision-making problem, we proposed the TDM framework in this paper by integrating symbolic planning with sequential decision-making operating on high-dimensional sensory input.
We also presented a TDM-based deep learning algorithm.
Deep learning architecture is used to learn low-level control policies for subtasks managed by high-level symbolic planning based on explicit symbolic knowledge. Both theoretical analysis and empirical studies on benchmark problems validate that our trust-score based algorithmic framework brings \textit{task-level interpretability} to deep reinforcement learning and \textit{improved data efficiency} induced by the symbolic-planning-learning framework of the agent.

There are several promising future work potentials in this research direction. For example, it would be intriguing to apply TDM in the human-computer interaction systems, allowing layperson to understand the system behavior and provide meaningful evaluation feedback to the machine. The second interesting direction is to apply the framework of TDM in multi-agent systems. Trust is important in interactions among different agents where the integrity of some agents is not always guaranteed. Interpretability may be used to help improve the behavior of the agents and lead to a better equilibrium.

\bibliographystyle{IEEEtranN}
\bibliography{bibliography}


\vspace{-20 mm}

\begin{IEEEbiography}[{\includegraphics[width=1in,height=1.25in,clip,keepaspectratio]{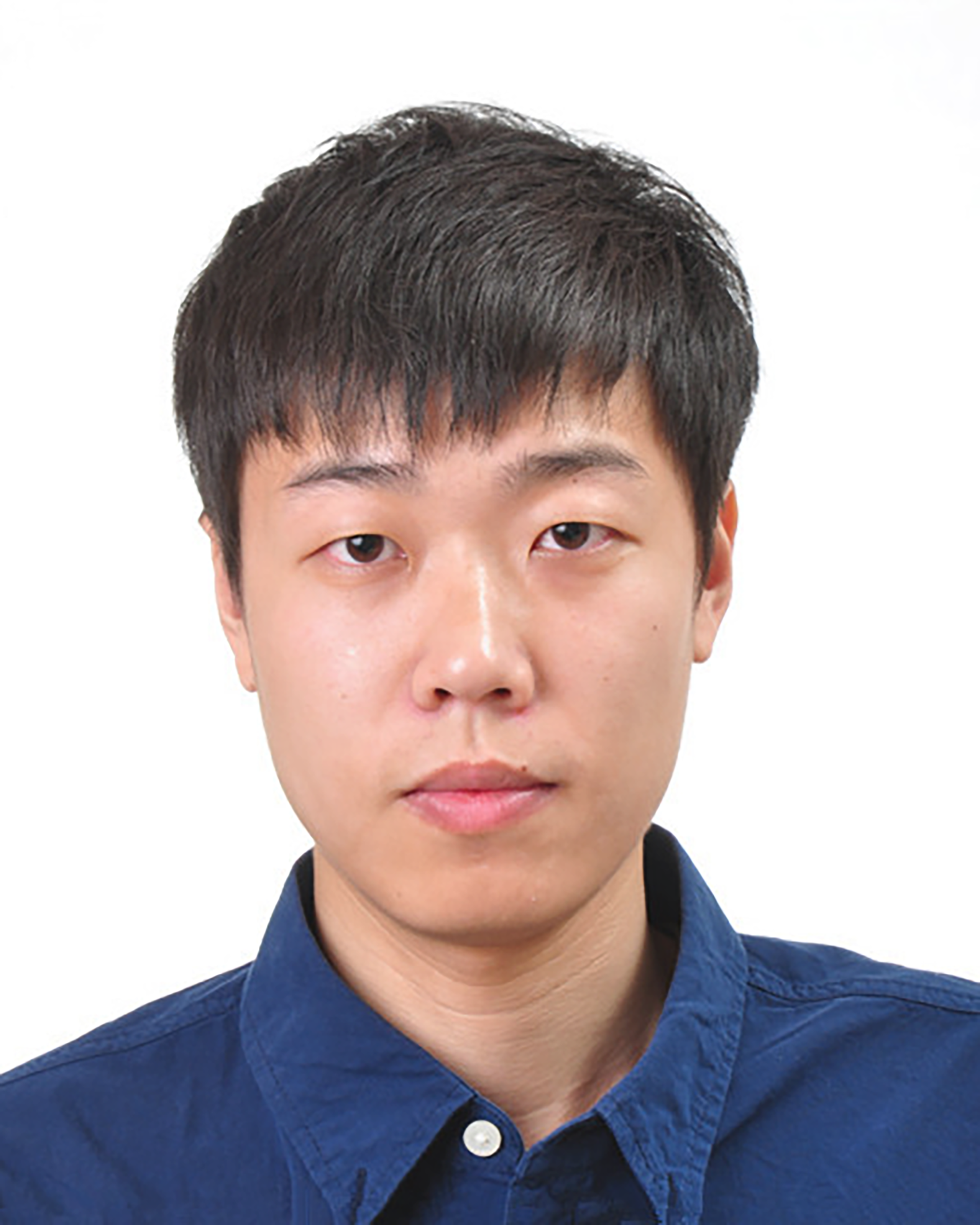}}]{Daoming Lyu}
received the B.S. degree in Electrical Engineering from Southwest University, Chongqing, China, in 2011, and the M.E. degree in Biomedical Engineering from Zhejiang University, Hangzhou, China, in 2015. He is currently pursuing the Ph.D. degree with the Department of Computer Science and Software Engineering, Auburn University, Auburn, AL, USA.
His current research interests include Reinforcement Learning, Neuro-Symbolic AI, Trustworthy Decision-making and Healthcare Informatics.
\end{IEEEbiography}

\vspace{-20 mm}

\begin{IEEEbiography}[{\includegraphics[width=1in,height=1.25in,clip,keepaspectratio]{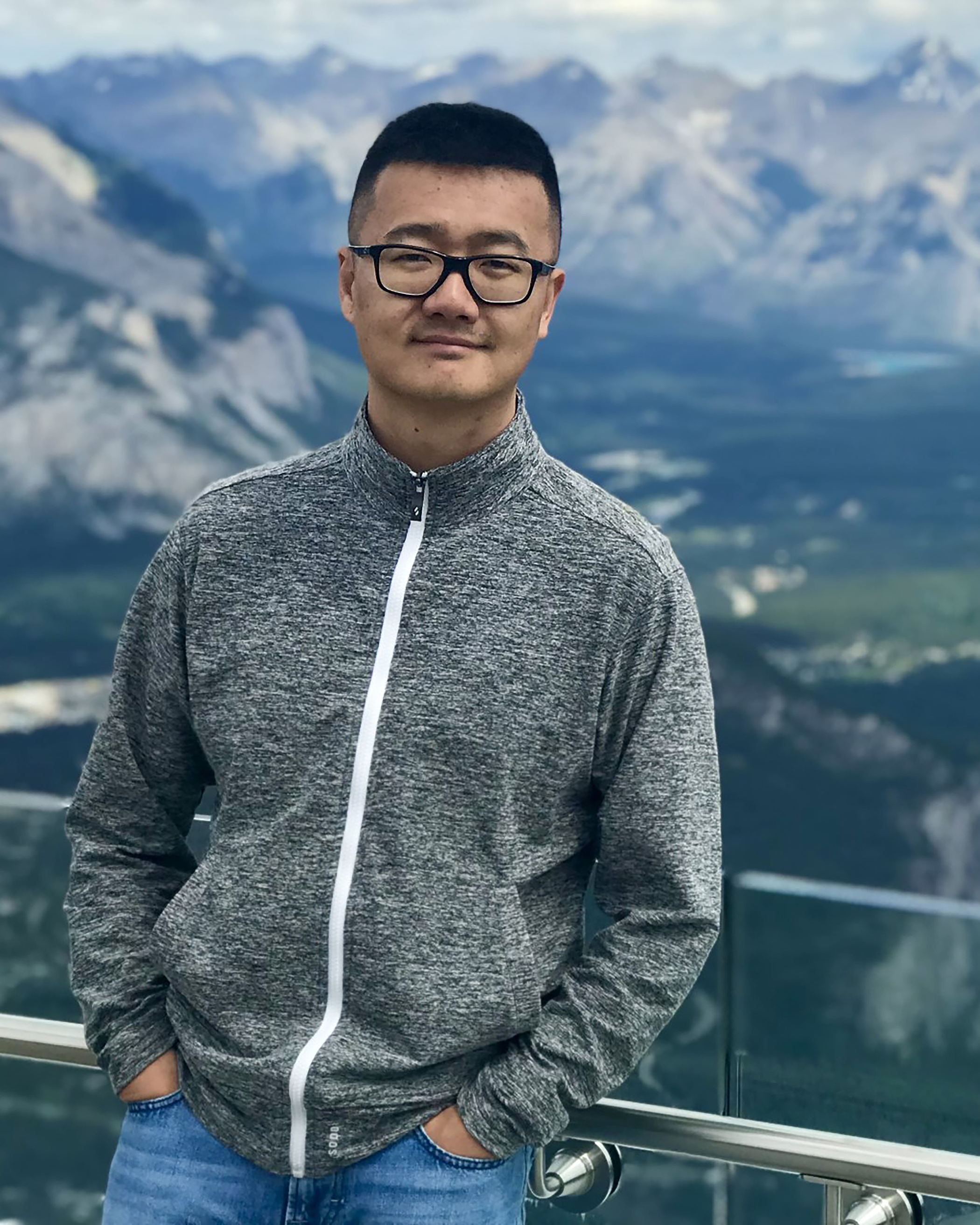}}]{Fangkai Yang}
obtained his Ph.D. from Department of Computer Science, University of Texas at Austin in 2014. His research focus is logic-based artificial intelligence, in particular, knowledge representation and reasoning, non-monotonic reasoning and answer set programming.
He is also interested in application of logic-based AI in autonomous systems, mobile robots and self-driving vehicles.
Dr. Yang is a senior scientist in NVIDIA focusing on behavior planning and prediction in self-driving vehicles.
\end{IEEEbiography}

\vspace{-20 mm}

\begin{IEEEbiography}[{\includegraphics[width=1in,height=1.25in,clip,keepaspectratio]{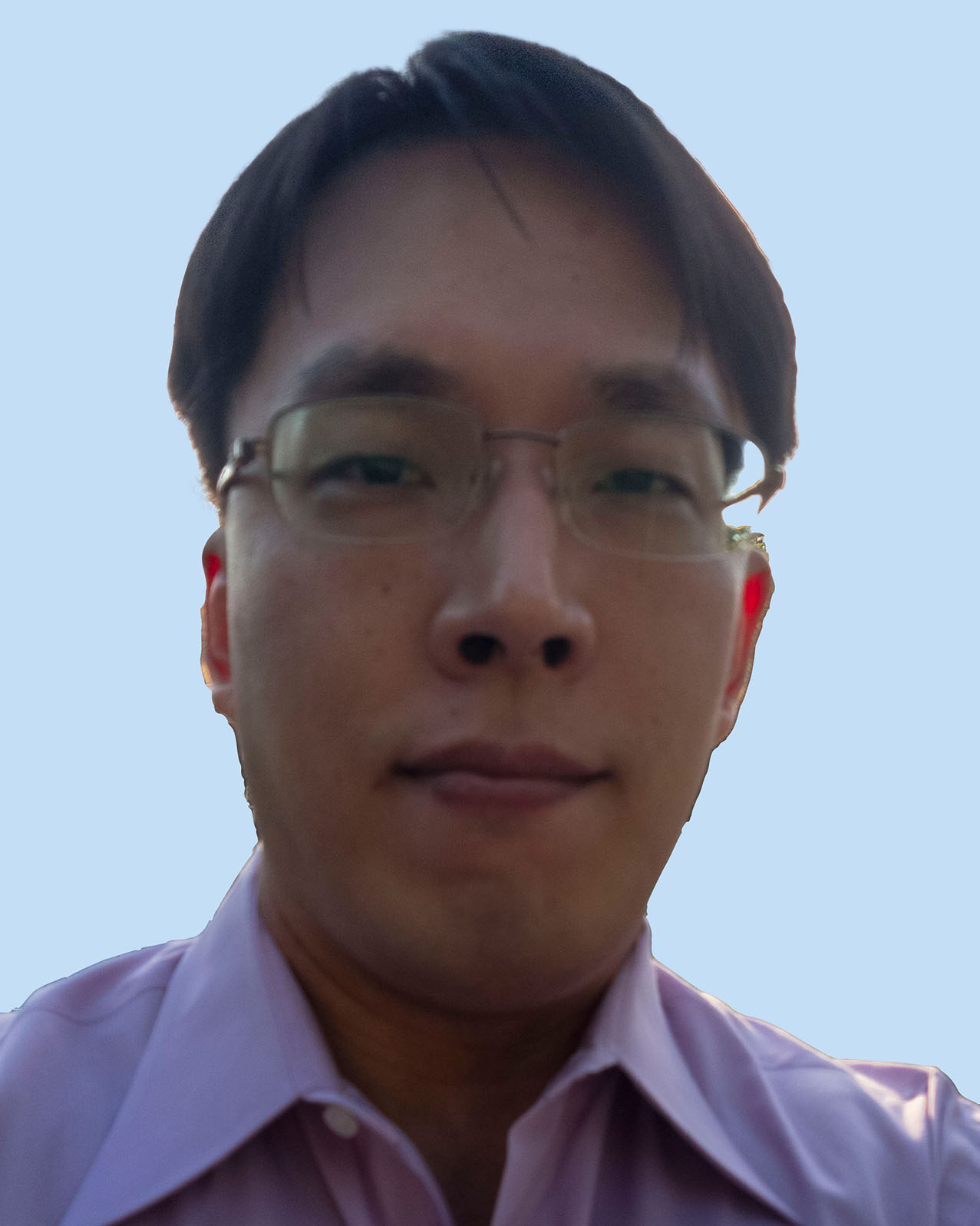}}]{Hugh Kwon}
received the B.S. and M.S. degrees in Computer Science from Columbus State University in 2009 and 2018, respectively. He is currently pursuing the Ph.D. with the Department of Computer Science and Software Engineering at Auburn University. His research interests include reinforcement learning, distributed learning, and safe learning.
\end{IEEEbiography}

\vspace{-20 mm}

\begin{IEEEbiography}[{\includegraphics[width=1in,height=1.25in,clip,keepaspectratio]{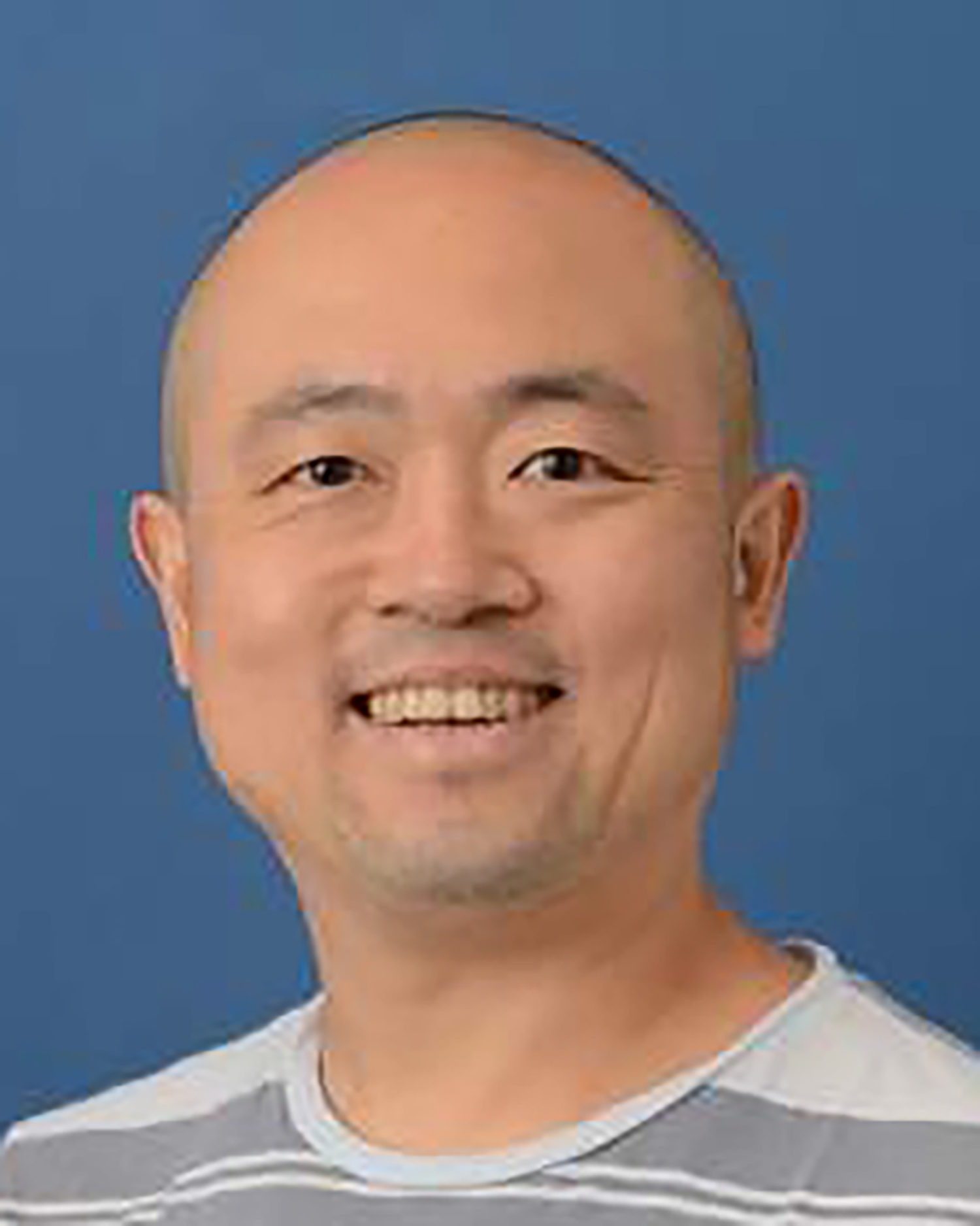}}]{Wen Dong}
is an Assistant Professor of Computer Science and Engineering with a joint appointment at the Institute of Sustainable Transportation and Logics at the State University of New York at Buffalo. His research focuses on developing machine learning and signal processing tools to study the dynamics of large social systems in vivo. He has a PhD degree from the MIT Media Laboratory.
\end{IEEEbiography}

\vspace{-20 mm}

\begin{IEEEbiography}[{\includegraphics[width=1in,height=1.25in,clip,keepaspectratio]{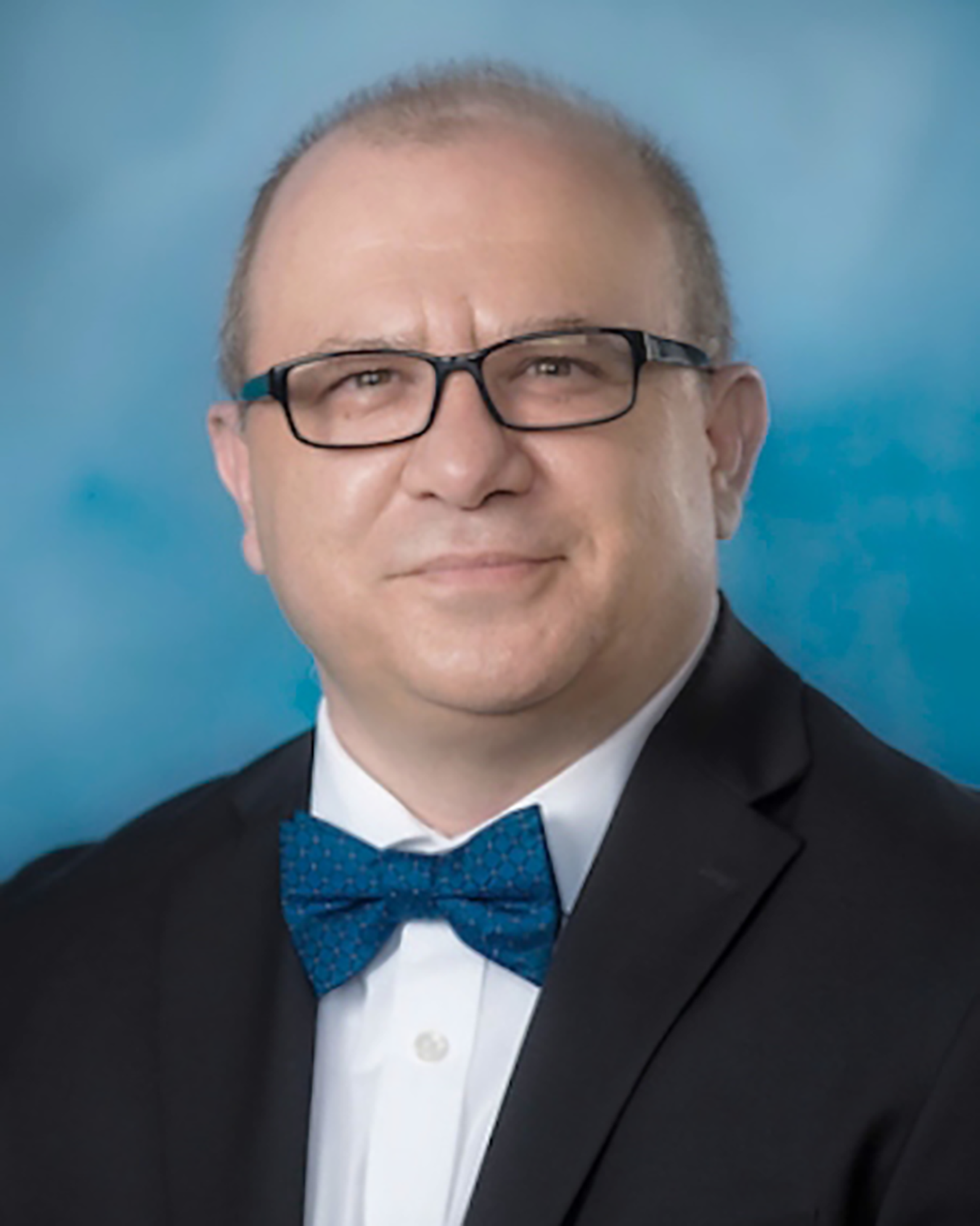}}]{Levent Yilmaz}
is  the Alumni Distinguished Professor of Computer Science and Software Engineering at Auburn University with a courtesy appointment in Industrial and Systems Engineering.
He holds M.S. and Ph.D. degrees in Computer Science from Virginia Tech.
His research interests are in theory and methodology of modeling and simulation, agent-directed simulation, cognitive systems, and model-driven science and engineering for complex adaptive systems.
\end{IEEEbiography}

\vspace{-20 mm}

\begin{IEEEbiography}[{\includegraphics[width=1in,height=1.25in,clip,keepaspectratio]{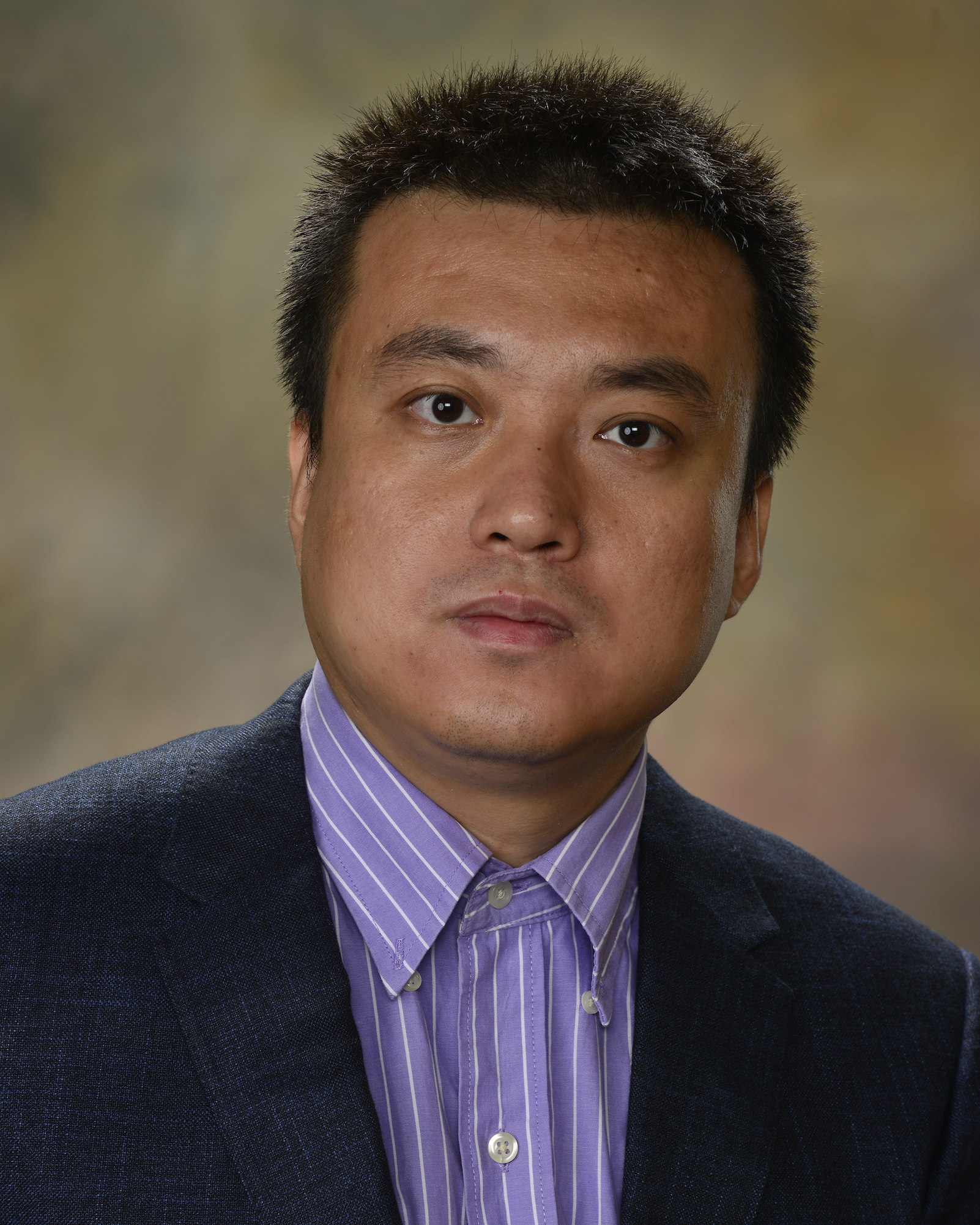}}]{Bo Liu} is a Tenure-Track Assistant Professor with the Department of Computer Science and Software Engineering, Auburn University, Auburn, AL, USA. He received the Ph.D. degree from the University of Massachusetts Amherst, Amherst, MA, USA, in 2015. He has over 30 publications on several notable venues. He is the recipient of the UAI'2015 Facebook best student paper award and the Amazon research award in 2018.  He is an Associate Editor of IEEE Transactions on Neural Networks and Learning Systems (IEEE-TNN), a senior member of IEEE, and a member of AAAI, ACM, and INFORMS.
\end{IEEEbiography}


\end{document}